\documentclass[11pt]{article}
\usepackage[T1]{fontenc}
\usepackage[utf8]{inputenc}
\usepackage{microtype}

\usepackage{booktabs}
\usepackage{amsmath}
\usepackage{mathtools}
\usepackage{newtxtext}
\usepackage{newtxmath}
\usepackage{tikz}
\usetikzlibrary{positioning, arrows.meta}
\definecolor{catmint}{HTML}{BFDFD2}
\definecolor{catteal}{HTML}{4098AC}
\definecolor{catgold}{HTML}{ECB66C}
\definecolor{catcoral}{HTML}{ED8D5A}
\definecolor{gradeGray}{HTML}{C9C9C9}
\usepackage{pgfplots}
\pgfplotsset{compat=1.18}
\usepackage{colortbl}  %
\usepackage{enumitem}
\setlist{leftmargin=*}  %
\newcommand{\hbfull}{\begin{tikzpicture}[baseline=-0.55ex]\fill[catteal] (0,0) circle (0.62ex);\end{tikzpicture}}
\newcommand{\hbhalf}{\begin{tikzpicture}[baseline=-0.55ex]\draw[catteal,line width=0.4pt] (0,0) circle (0.62ex);\fill[catteal] (0,0) -- (0,0.62ex) arc (90:-90:0.62ex) -- cycle;\end{tikzpicture}}
\newcommand{\hbnone}{\begin{tikzpicture}[baseline=-0.55ex]\draw[catteal!55,line width=0.4pt] (0,0) circle (0.62ex);\end{tikzpicture}}

\newif\ifarrreview
\arrreviewfalse         %

\ifarrreview
  \usepackage[review]{acl}
  \makeatletter
  \@ifpackagelater{lineno}{2026/02/11}{%
    \IfFormatAtLeastTF{2025-06-01}{%
      \RemoveFromHook{build/column/before}[lineno]%
      \AddToHook{build/column/after}{%
        \setbox\@outputbox\vbox{%
          \boxmaxdepth\@maxdepth
          \protected@write\@auxout{}{%
            \string\def\string\@LN@column{\if@firstcolumn1\else2\fi}}%
          \box\@outputbox}}%
    }{}%
  }{}
  \makeatother
\else
  \usepackage[preprint]{acl}
\fi
\newcommand{\Description}[1]{}
\usepackage{subcaption}

\newcommand{\FaultEarlyPct}{57}     %
\newcommand{\FaultCentralPct}{54}   %

\newcommand{\NObsCorpora}{six}      %
\newcommand{\TauFlat}{0.619}        %
\newcommand{\OhFlat}{0.698}         %
\newcommand{\TauTwoFlat}{0.722}     %
\newcommand{\WebFlat}{0.765}        %
\newcommand{\SweGymDepGain}{+0.142} %
\newcommand{\TauShapeGain}{+0.046}  %
\newcommand{\SweShapeGain}{+0.036}  %
\newcommand{\SweGymShapeGain}{+0.064} %

\newcommand{\secref}[1]{\S\ref{#1}}
\newcommand{\appref}[1]{Appx.~\ref{#1}}
\newcommand{\grade}{\textsc{Grade}}  %
\newcommand{\MaxDepSizeCorr}{0.96}     %
\newcommand{\MaxInfDepSizeCorr}{0.99}  %

\newcommand{\SweGymTrBoth}{0.590}

\newcommand{\TauFlatDens}{0.620}   %

\newcommand{\TauFlatDep}{0.665}

\newcommand{\NTwoTrMeanGap}{+0.028}   %
\newcommand{\NTwoTrAbove}{5}          %
\newcommand{\NZeroTrAbove}{3}         %
\newcommand{\NTwoTrSeparates}{three}  %
\newcommand{\NTwoTrInverts}{two}      %
\newcommand{\NTwoTrSweGymP}{0.005}    %
\newcommand{\NTwoTrTauWebP}{0.020}    %
\newcommand{\NTwoTrDraws}{200}

\newcommand{\NTwoCiSweGym}{[+0.086, +0.133]}
\newcommand{\NTwoCiMean}{[+0.021, +0.036]}

\newcommand{\NLadderRuns}{3,114}
\newcommand{\NZeroEdgeRatio}{5.3}   %

\newcommand{\NTwoTrResetGap}{+0.029}

\newcommand{\FamFamilies}{three}     %
\newcommand{\FamNewTargets}{five}    %
\newcommand{\FamSweGymDelta}{-0.036} %
\newcommand{\FamAwareSweGym}{0.343}  %
\newcommand{\FamSweGymDrop}{-0.247}  %
\newcommand{\FamSOne}{-0.016}        %
\newcommand{\FamSOneCI}{[-0.027, -0.004]}
\newcommand{\FamSTwo}{-0.013}        %
\newcommand{\FamSTwoCI}{[-0.021, -0.006]}
\newcommand{\FamVerifyN}{222}        %
\newcommand{\FamVerifyMax}{6.5\times10^{-16}}  %

\begin{document}

\title{\grade: Graph Representation of
LLM Agent Dependency and Execution}

\ifarrreview
  \author{Anonymous ACL submission}
\else
  \author{Yue Zhao \\
    University of Southern California \\
    \texttt{yue.z@usc.edu}}
\fi

\maketitle

\begin{abstract}
Traces record what LLM agents did, leaving implicit what each step relied on. \grade{} holds both
in one graph; six published graph views are its projections. Execution edges come free; each costly
dependency edge is graded as observed in the trace, declared by added logging, or inferred under an
assumption, a schema we test one grade at a time. We recover the dependency layer from \NObsCorpora{} tool-use, coding, and web corpora
and measure its failure-prediction gain over run size. First, over a linear run-size baseline, it
improves prediction on three corpora, but the gain depends on the learner and task sampling.
On held-out corpora the layer stays above chance where run size does not, under one learner, with
the same task-sampling caveat. Second, structure can pass for signal it does not carry: a layer inferred
under full history restates run size. Within a corpus, a preregistered test does not separate the
recovered layer from a degree-matched counterfeit that rewires each step's dependencies but keeps
their number. The recovered layer's largest transfer lead over it reverses when the held-out
corpus's task family is withheld. Before crediting structure, grade each edge, keep a run-size
baseline, and score degree-matched attachment controls.
\end{abstract}

\section{Introduction}
\label{sec:intro}

\textbf{LLM agents fail for what they relied on, but the record shows only what they did.} A booking
agent reads a price, holds a reservation, then confirms it; the confirmation is valid only while the
price is current. If the price goes stale in between, the run fails, yet every step ran in order. Such
unchecked inputs recur in agent failures~\cite{cemri2025mast}.

\begin{figure*}[!t]
\captionsetup{skip=4.5pt}
\centering
\includegraphics[width=\textwidth]{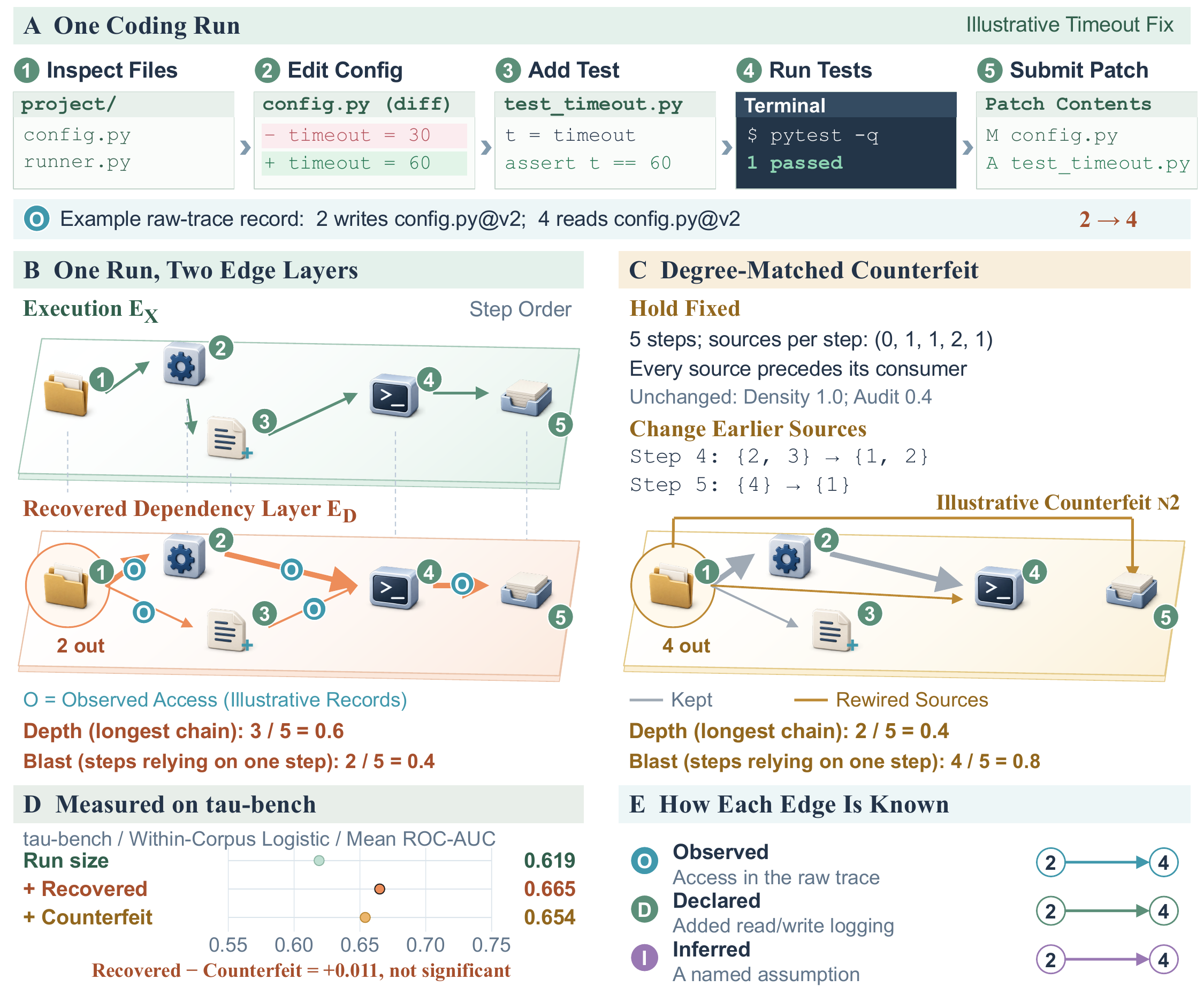}
\caption{\textbf{One run, two layers, graded edges.} (A) A coding run and a raw-trace record. (B) The same run as two layers on shared nodes: execution (step order) and dependency (what each step used). (C) The degree-matched counterfeit \textsc{n2} rewires which earlier steps each step depends on but keeps their number, so density and audit stay fixed. (D) Within-corpus ROC-AUC on tau-bench: the recovered layer adds to the run-size baseline and scores slightly above the counterfeit, a gap that does not survive correction (Table~\ref{tab:ladder}, Figure~\ref{fig:ladderci}). (E) Dependency edges are graded observed (in the trace), declared (added logging), or inferred (a named assumption).}
\Description{A five-step coding run drawn as an execution layer and a dependency layer on shared nodes, a degree-matched counterfeit that changes which earlier step each edge attaches to, scores for tau-bench, and three grades for how a dependency edge is known.}
\label{fig:twolayer}
\vspace{-10pt}
\end{figure*}

\textit{Coordination failures} live in the execution: which agent
acted, how control passed, whether a handoff dropped a
result~\cite{cemri2025mast,zhang2025whoandwhen}. 
\textit{Reliance failures} live in the dependency structure (Figure~\ref{fig:twolayer}B), which no trace states. A step
used state an earlier step produced or held, now stale, overwritten, or wrong from the start.

A second problem travels with it: \textit{there is no shared way to draw such runs}. LLM agents span
tool use, coding, and multi-agent collaboration~\cite{xi2023agentsurvey}, each drawn its own way:
an orchestration graph~\cite{zhuge2024gptswarm,hong2024metagpt}, a reason-act
trace~\cite{yao2023react}, or a sequence of edits~\cite{yang2024sweagent,jimenez2024swebench}.
Analysis is rebuilt per setting. Both share one cause: \textbf{common representations of
agent runs \textit{center} execution and \textit{leave} dependency implicit}. Dependency is the
harder half to obtain, from trace content, instrumentation, or an assumption; once recorded, it
would carry analysis across settings. We propose \textbf{\grade{} (\underline{G}raph
\underline{R}epresentation of LLM \underline{A}gent \underline{D}ependency and
\underline{E}xecution)}, a unified representation for agent runs (\secref{sec:representation}).

Program analysis has represented dependency as a graph since the 1980s: program dependence
graphs~\cite{ferrante1987pdg}, static interprocedural slicing~\cite{horwitz1990interprocedural},
dynamic slicing~\cite{korel1988dynamic}, and dynamic dependence graphs over executed statement
occurrences~\cite{agrawal1990dynamic}. \grade{} takes that relation into agent traces. There the
evidence behind an edge varies by corpus (Figure~\ref{fig:twolayer}A), so each dependency edge is
graded by how it is known.

\textit{What does the costly dependency layer add over the free execution one}? Run size (the
counts of steps, tool calls, decisions, and agents) comes free from the execution layer. We therefore
measure the \emph{gain over run size}: how much the dependency layer improves failure prediction over
a linear model of run size (\secref{sec:twomodes}). This paper makes three contributions.

\begin{itemize}[topsep=4pt,itemsep=3pt,parsep=0pt]
\item \textbf{A Unified Representation.} \grade{} records a run as one graph with execution and
dependency layers on the same steps, each dependency edge graded by how it is known
(\secref{sec:representation}, Figure~\ref{fig:twolayer}). Six published graph views of agent runs, from
reason-act traces to lineage graphs, are its projections under the conditions in
Table~\ref{tab:subsumption}. The per-edge grade is a proposed schema: our experiments test each grade
only as a whole layer.

\item \textbf{Empirical Findings.} Over a linear run-size baseline, the recovered dependency layer
improves failure prediction on three of the six corpora, but the gain depends on the learner and task sampling
(\secref{sec:signal}). On held-out corpora it stays above chance where run size does not, under one
learner, with the same task-sampling caveat (\secref{sec:unification}). Graph structure can also
pass for signal it does not carry: a layer inferred under full history restates run size
(\secref{subsec:degenerate}). Within a corpus, a preregistered test does not separate the recovered
layer from a degree-matched counterfeit, which rewires each step's dependencies but keeps their
number (\secref{subsec:ladder}). The recovered layer's largest transfer lead over it reverses when
the held-out corpus's task family is withheld (\appref{app:familytransfer}).

\item \textbf{A Reusable Evaluation Protocol.} Before graph structure gets credit, the protocol
grades each edge, keeps a run-size baseline, and scores degree-matched counterfeits within a corpus
and in transfer (\secref{sec:twomodes}, \secref{subsec:ladder}, \secref{subsec:gate}). The
anonymized supplementary material holds its code, the loaders that recover each corpus's dependency
layer, and a data card per corpus (\appref{app:corpuscard}).\ifarrreview\else\footnote{Available at
\url{https://github.com/yzhao062/grade}.}\fi

\end{itemize}

\begin{table*}[t]
\centering\footnotesize
\setlength{\tabcolsep}{3pt}
\captionsetup{skip=4.5pt}
\renewcommand{\arraystretch}{1.0}
\caption{\textbf{Six published graph views as projections of the two-layer graph.} Discs mark what
each view keeps of the execution layer, the dependency layer and the per-edge grade:
\protect\hbfull{} kept, \protect\hbhalf{} partial, \protect\hbnone{} absent. The last column is the
condition under which the view lines up with \grade{}; it is not a recovery theorem, and it says
nothing about how a model fitted on one view would behave (each view is described in
\appref{app:formal}).}
\label{tab:subsumption}
\begin{tabular}{@{}%
>{\raggedright\arraybackslash}p{0.235\textwidth}%
ccc%
 !{\color{gradeGray}\vrule width 0.4pt}%
>{\raggedright\arraybackslash}p{0.585\textwidth}@{}}
\toprule
Graph view & Exec. & Dep. & Grade & Lines up with the two-layer graph when \\
\midrule
Reason-act / tool trace~\cite{yao2023react} & \hbfull & \hbhalf & \hbnone &
A reliance edge joins two steps only where both touch the same named resource \\
MetaGPT orchestration~\cite{hong2024metagpt} & \hbfull & \hbhalf & \hbnone &
Subscription fixes the artifact kind a role reads, never the revision a step used \\
GPTSwarm graph~\cite{zhuge2024gptswarm} & \hbfull & \hbhalf & \hbnone &
Routines that ignore their input, a gap the view itself names, stay apart (\appref{app:formal}) \\
Computation graph~\cite{acg2026survey} & \hbfull & \hbhalf & \hbnone &
An emitter types the edges; the view has no typing function of its own \\
Data-lineage / PROV~\cite{moreau2013provdm} & \hbnone & \hbfull & \hbnone &
Steps that ran in sequence and exchanged nothing get no edge~\cite{cheney2013provconstraints} \\
Agent anomaly graph~\cite{he2025sentinelagent,zhou2025guardian} & \hbfull & \hbhalf & \hbhalf &
State dependency stays partial; SentinelAgent pairs a declared graph with an observed one \\
\bottomrule
\end{tabular}
\vspace{-10pt}
\end{table*}

\section{A Unified Graph Representation for LLM Agents}
\label{sec:representation}

\subsection{Execution Edges and Graded Dependency Edges}
\label{subsec:layers}

\grade{} records a run as one graph with two edge layers on the same nodes
(Figure~\ref{fig:twolayer}B). The nodes are agents, decisions, tool calls, and external
resources, the state a decision reads or writes (a database row, a source file). Execution edges
record control: an agent \emph{emits} its decision and tool nodes, and control \emph{hands off}
from one step to the next. Dependency edges record reliance: a step \emph{depends on} an earlier
node whose state it used, \emph{reads} a resource, or \emph{writes} one. Which earlier step a
dependency edge points to is its \emph{attachment}. Multi-agent, tool-use, coding, and web traces
build this graph the same way.

The two layers differ in cost. Execution edges are free, because a trace records which agent
acted and in what order. No trace states what a step relied on, so each dependency edge is
graded by how it is known (Figure~\ref{fig:twolayer}E). An edge is \emph{observed} when the trace
shows the access it records, as when a coding step edits a named file or a database step reads a
named row. It is \emph{declared} when added logging records the reads and writes a raw trace
omits. Where no access is logged, an edge is \emph{inferred} under a named assumption, the weakest
being full history: each step used every earlier step (\appref{app:formal}). Grades attach to
single edges, so one graph can mix observed edges with inferred ones where the trace fell silent.
We evaluate each grade only as a whole layer; mixed-grade graphs are untested.

\subsection{The Formal Graph and Its Projections}
\label{subsec:formal}

In \grade{}, a run is a directed, typed, temporal multigraph $G = (V, E_X, E_D, \tau, t, \sigma)$:
nodes $V$ with types $\tau$ and times $t$, an execution layer $E_X$ read deterministically off the
trace, a dependency layer $E_D$ recovered from logged accesses or inferred under a named
assumption, and a map $\sigma$ that grades each dependency edge \textsf{observed},
\textsf{declared}, or \textsf{inferred}. The full tuple and recovery maps are in
\appref{app:formal}.

\label{subsec:projections}

Four operations produce the fragments of Table~\ref{tab:subsumption}: deleting node types, deleting
or merging an edge layer, contracting steps onto the agent that emitted them, and deleting the
per-edge grade. Reasoning graphs in the Graph-of-Thoughts family~\cite{besta2024got}, Tree of
Thoughts included, sit outside the table. Their vertices are units of information rather than
run steps, and they can be deleted mid-run, which no projection of a step graph reproduces.

\section{What Dependency Adds over Run Size}
\label{sec:twomodes}

The execution layer gives run size for free, so we credit the costly dependency layer only with
what it adds to failure prediction over run size. Run size is four counts from the trace: steps,
tool calls, decisions, and agents. A linear logistic model on these counts is the run-size
baseline. It leaves out per-step execution features, such as the acting agent's share of
decisions, activity rank and handoff degrees, which an ancillary study scores on Who\&When
(\appref{app:loc}). The dependency features are four size-normalized summaries of the
dependency layer (Figure~\ref{fig:twolayer}B, C; formulas in \appref{app:within}). They are density
(edges per step), depth (the longest dependency chain), blast (the largest share of the run that
relies directly on one step) and audit (the largest share of the run that one step relies on). Their
\emph{gain over run size} is how much cross-validated ROC-AUC rises when they join the run-size
baseline on identical splits (Figure~\ref{fig:twolayer}D):
\[
  \begin{aligned}
    \Delta_{\mathrm{dep}}
    \;=\;
    &\mathrm{AUC}\!\left(\text{run size} + \text{dependency}\right)\\
    &{}-\mathrm{AUC}\!\left(\text{run size}\right).
  \end{aligned}
\]

A positive gain shows what the dependency features add to a linear model of four counts, which
is weaker than information beyond run size. Under this learner, a nonlinear function of run size
can raise the AUC while carrying nothing the counts did not already hold
(\secref{sec:misread} gives a case). Separating the gain from such a re-encoding takes two
further tests: a preregistered test against counterfeit layers that rewire which earlier step
each edge points to (\secref{subsec:ladder}), and transfer to a held-out corpus
(\secref{sec:unification}). The gain may come out near zero where run size already separates
failures. Two precautions keep it honest: run-length normalization and the nested comparison on
identical splits (\appref{app:object}).

\section{Does Dependency Predict Failure?}
\label{sec:dependency}

\subsection{Prediction Within a Corpus}
\label{sec:signal}

\textbf{Corpora and model.} We first ask whether the recovered dependency layer predicts failure
within each corpus. Every dependency edge in the \NObsCorpora{} corpora is observed. Two are
database tool-use corpora scored against a gold final state (tau-bench~\cite{yao2024taubench},
tau2-bench~\cite{barres2025tau2bench}). Three are software-engineering corpora scored on issue
resolution (SWE-agent~\cite{yang2024sweagent}, SWE-Gym~\cite{pan2024swegym},
OpenHands~\cite{wang2024openhands}), and one is a web-navigation corpus
(AgentRewardBench~\cite{lu2025agentrewardbench}). A standardized logistic regression predicts failure; we score it by ROC-AUC under five-fold cross-validation over five
seeds. Normalizing the features by run length does not free them of run size, so each gain compares
models with and without the dependency features on identical splits. A gain clears zero when its
95\% interval across the five split seeds excludes zero; this split-seed interval measures
sensitivity to the split and says nothing about task sampling (\appref{app:within}).

\textbf{One grade, three attachment rules.} On tau-bench a consequential write attaches to every
prior database read, an upper bound because the trace does not say which read the write used. In the file and web corpora an access attaches to the most recent earlier access
to the same resource. On tau2-bench a step attaches to the most recent earlier step that produced an
identifier it consumes (\appref{app:corpuscard}). An observed edge records shared access to a
resource, not proven reliance.

\begin{table}[t]
\centering\scriptsize
\setlength{\tabcolsep}{2pt}
\caption{Within-corpus gain $\Delta$ in ROC-AUC when the dependency features join a baseline. Logistic model, run-level folds (\appref{app:within}). \textsc{flat} is the run-size baseline; the composition baseline adds counts of reads, writes and resources (\appref{app:c1}). The task interval resamples tasks ($2000$ draws) for the gain on pooled out-of-fold predictions, which can differ from $\Delta$. A dash marks a corpus with too few task clusters for that interval.}
\label{tab:within}
\begin{tabular}{@{}lcccc@{}}
\toprule
 & \multicolumn{2}{c}{Over \textsc{flat} baseline} & \multicolumn{2}{c}{Over composition baseline} \\
\cmidrule(lr){2-3}\cmidrule(lr){4-5}
Corpus & $\Delta$ & Task 95\% CI & $\Delta$ & Task 95\% CI \\
\midrule
tau-bench & $+0.046$ & $[-0.005, +0.100]$ & $+0.018$ & $[-0.008, +0.043]$ \\
SWE-agent & $+0.085$ & --- & $+0.065$ & --- \\
SWE-Gym & $+0.142$ & $[+0.084, +0.217]$ & $+0.012$ & $[-0.017, +0.036]$ \\
OpenHands & $-0.004$ & $[-0.015, +0.004]$ & $-0.002$ & $[-0.004, -0.001]$ \\
tau2-bench & $-0.015$ & $[-0.028, -0.003]$ & $-0.016$ & $[-0.029, -0.001]$ \\
web & $-0.009$ & $[-0.024, +0.005]$ & $-0.002$ & $[-0.016, +0.013]$ \\
\bottomrule
\end{tabular}
\end{table}

\subsubsection{The Gain over Run Size in Each Corpus}

\textbf{Over the linear run-size baseline, the dependency features add a gain on three corpora and
none on the other three.} We order the corpora by how well that baseline predicts failure
(Table~\ref{tab:ladder}), from the size-weak corpora tau-bench, SWE-agent and SWE-Gym to the
size-strong corpora OpenHands, tau2-bench and web. Each size-weak corpus shows a gain in all five
seeds, with a split-seed interval clear of zero (Table~\ref{tab:within}, Figure~\ref{fig:keystone}).
On the size-strong corpora the gain is consistent with zero on OpenHands and small and negative in every seed on tau2-bench and web. Because
the ordering and each gain share the run-size baseline, the two are arithmetically coupled. A
permutation check shows that this design cannot separate a dependency effect from that coupling
(\appref{app:ordering}). Each corpus's gain keeps its own sign and interval, computed without
reference to the ordering.

\begin{figure}[t]
\centering
\includegraphics[width=\columnwidth]{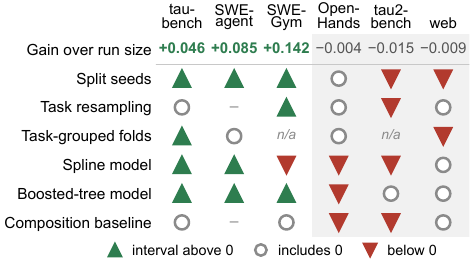}
\caption{\textbf{Which checks the within-corpus gain survives.} Top row: the gain over run size
(Table~\ref{tab:within}). Each other row: where that check's 95\% interval lies (task resampling for the
task-resampling and composition rows, split seeds elsewhere; \appref{app:c1}, \appref{app:c2}). n/a:
one run per task; dash: too few task clusters.}
\Description{A grid with one column per corpus and one row per robustness check. Upward triangles
mark an interval above zero, circles an interval that includes zero, and downward triangles an
interval below zero. The three corpora with a positive lift keep it under split seeds, and each
loses it under at least one other check.}
\label{fig:liftgrid}
\end{figure}

\textbf{The gain depends on the baseline and the learner} (Figure~\ref{fig:liftgrid}). Over a
composition baseline that also counts reads, writes and resources (Table~\ref{tab:within}), the three
gains shrink but stay positive in every seed. None keeps an interval above zero when tasks are
resampled. With $1/n$ and $\log n$ added to the four counts, only tau-bench's
increment keeps a seed-block interval above zero under both run-level and task-grouped folds
($+0.046$ and $+0.028$), and its task-level intervals include zero; SWE-Gym's falls from
\SweGymDepGain{} to $-0.006$ (\appref{app:flexsize}). Under a spline model SWE-Gym's gain is negative, and under a boosted tree it is positive.
OpenHands and tau2-bench are negative under all three learners (\appref{app:c1}).

\textbf{So does task sampling.} When tasks are resampled instead of runs, only SWE-Gym's gain stays
above zero, and SWE-agent has too few tasks with a resolved run to tell (Table~\ref{tab:within}).
Grouping folds by task turns SWE-agent's $+0.085$ into $-0.011$ and cuts tau-bench's to $+0.028$. SWE-Gym's one run per task makes grouping there vacuous (\appref{app:c2}).

\textbf{The gain passes two checks, and neither ties it to the attachment.} It persists when the
model also sees how often a run revisits state, so it does not restate revisit volume. On the
size-strong corpora the dependency features alone still predict failure, so their overlap with run
size is shared signal rather than an empty layer (\appref{app:within}).

\subsection{A Preregistered Test Against Counterfeit Layers}
\label{subsec:ladder}

\textbf{A gain over run size does not show that the recovered attachment is the reason.} The
logistic model reads four summaries of the dependency layer, and a counterfeit layer of the same
shape would supply them too. We therefore build counterfeit layers that keep the steps and their
order and change only \emph{which} earlier step each edge points to.

\textbf{Four counterfeit layers compete with the recovered dependency layer.} The reading-free
chain links each step to the step just before it and reads nothing from the log, so in total it carries
more edges than the recovered layer. Each of the other three keeps the recovered edge count on every run.
Of these, the uniform counterfeit places those edges uniformly among earlier-to-later step pairs. The
degree-matched counterfeit keeps how many earlier steps each step depends on but rewires which ones
(Figure~\ref{fig:twolayer}C). Last, the recency counterfeit keeps those counts but attaches to the most
recent earlier steps (constructions in \appref{app:ladder}). Such counterfeits follow network null
models \cite{vasa2022nullmodels,maslov2002specificity}; scoring a model against a matched
counterfeit it should not fit follows \citet{hewitt2019control}.

\begin{figure}[t]
\centering
\captionsetup{skip=4.5pt}
\includegraphics[width=\columnwidth]{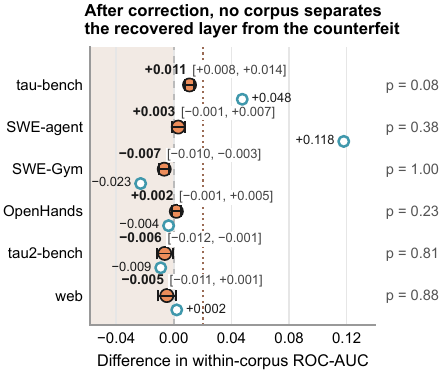}
\vspace{-10pt}
\Description{A horizontal interval plot, one row per corpus, of the difference between the recovered
attachment and the degree-matched control, with the permutation p printed at the right and a hollow
marker for the chain contrast against the H2 threshold line.}
\caption{\textbf{Within a corpus, the recovered layer does not separate from the degree-matched counterfeit.} Coral discs: $d$ (Table~\ref{tab:ladder}) with seed-block $95\%$ intervals over five seeds; right column: permutation $p$ against degree-matched draws, $B=25$; hollow markers: \textsc{aware} minus \textsc{n0}; dotted line: the second prediction's $+0.02$ threshold (H2). No $p$ survives Holm across \NObsCorpora{} corpora.}
\label{fig:ladderci}
\vspace{-10pt}
\end{figure}

\textbf{Neither preregistered prediction held.} First, the recovered layer would beat the
degree-matched counterfeit; after correction for multiple tests, no corpus separates the two
(Figure~\ref{fig:ladderci}). Second, it would beat the reading-free chain by more than $0.02$ in
ROC-AUC on most corpora. It does so on two of \NObsCorpora{}.

\textbf{The reading-free chain beats the recovered layer on three corpora}, SWE-Gym, OpenHands and
tau2-bench (Table~\ref{tab:ladder}), and loses on the other three. Concurrent work reports the
opposite against a weaker counterfeit: RUPA finds its graph beats a random one \cite{ma2026rupa}.
The two results are compatible, since the strength of the counterfeit decides the answer. Our own
results show this on tau-bench, where the recovered layer's lead shrinks from the uniform
counterfeit to the degree-matched one (Table~\ref{tab:ladder}). A relation-aware graph network,
scored against a counterfeit of the uniform kind, also favors the recovered layer
(\appref{app:gin}).

\begin{table}[t]
\centering\footnotesize
\setlength{\tabcolsep}{2pt}
\captionsetup{skip=5pt}
\caption{\textbf{The control ladder: within-corpus ROC-AUC as only the attachment changes.} \textsc{flat} is the run-size baseline, \textsc{aware} the recovered layer, and \textsc{n0} to \textsc{n3} the reading-free chain and the uniform, degree-matched and recency counterfeits, in order of increasing strength. \textsc{n1} to \textsc{n3} hold the edge count fixed as well; \textsc{n0} does not, and
carries \NZeroEdgeRatio{} times the observed count. $d = \textsc{aware} - \textsc{n2}$ is the primary comparison. Neither preregistered prediction held:
$d$ has a six-corpus mean of $-0.0003$, and \textsc{aware} clears \textsc{n0} by more than $+0.02$
on two corpora of \NObsCorpora{}. Per-seed values: \appref{app:ladder}.}
\label{tab:ladder}
\resizebox{\columnwidth}{!}{%
\begin{tabular}{@{}lrccccccr@{}}
\toprule
Corpus & $n$ & \textsc{flat} & \textsc{aware} & \textsc{n0} & \textsc{n1} & \textsc{n2} & \textsc{n3} & $d$ \\
\midrule
tau-bench  & 660 & 0.619 & 0.665 & 0.618 & 0.614 & 0.654 & 0.651 & $+0.011$ \\
SWE-agent  & 600 & 0.628 & 0.713 & 0.595 & 0.698 & 0.710 & 0.721 & $+0.003$ \\
SWE-Gym    & 376 & 0.663 & 0.804 & 0.827 & 0.814 & 0.811 & 0.806 & $-0.007$ \\
OpenHands  & 600 & 0.698 & 0.695 & 0.698 & 0.689 & 0.693 & 0.696 & $+0.002$ \\
tau2-bench & 278 & 0.722 & 0.706 & 0.715 & 0.709 & 0.713 & 0.714 & $-0.006$ \\
web        & 600 & 0.765 & 0.756 & 0.754 & 0.759 & 0.761 & 0.758 & $-0.005$ \\
\bottomrule
\end{tabular}%
}
\vspace{-2pt}
\end{table}

\textbf{The null result is narrow.} It covers the evaluated extraction, features, logistic model,
corpora and degree-matched counterfeit, and we report it because it was preregistered. The result
does not show that the attachment or the grade is inert, and it does not carry to other projections, features or
learners. Two of the four features, density and audit, cannot move under the degree-matched or
recency counterfeit, because each step's number of earlier sources, which both keep, fixes them
(Figure~\ref{fig:twolayer}C). That bounds what this test could detect.

\subsection{Transfer to a Held-Out Corpus}
\label{sec:unification}

\textbf{We train on five corpora and test on the held-out sixth.} The held-out corpus rotates through all \NObsCorpora{}. Each rotation fits the logistic model of \secref{sec:signal} to three feature sets: run size alone, the dependency features alone, and run size plus the dependency features. Pairwise transfer is in \appref{app:misread}. Training on five corpora at once does not require a feature to predict failure in each of them. A score therefore describes portability under one learner rather than a mechanism shared across corpora.

\begin{figure}[t]
\centering
\includegraphics[width=\columnwidth]{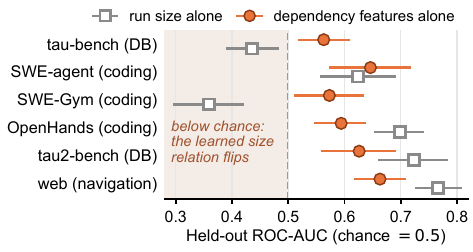}
\caption{\textbf{In transfer, run size flips on two corpora; the dependency features do not.} One logistic model, held-out bootstrap 95\% intervals (Table~\ref{tab:transferci}).}
\Description{Dot plot with one row per held-out corpus. Run size falls below the 0.5 chance line on
tau-bench and SWE-Gym; the dependency features alone stay above it for every held-out corpus.}
\label{fig:transferflip}
\end{figure}

\textbf{Under the logistic model the dependency features stay above chance on every held-out corpus;
run size does not.} Run size transfers to four held-out corpora but falls below chance on tau-bench
and SWE-Gym, where the relation between run size and failure learned elsewhere flips
(Figure~\ref{fig:transferflip}). The dependency features alone never fall below chance, and neither
does run size plus the dependency features. When held-out runs are resampled, every interval of
these two feature sets lies above chance, while both run-size inversions lie below it
(\appref{app:transferci}).

\textbf{The pattern weakens under task resampling and other learners.} When held-out tasks are
resampled instead of runs, both dependency intervals on tau-bench include chance. The paired gain of
the dependency features alone over run size stays above zero there (Table~\ref{tab:transfertask}).
SWE-agent has too few tasks with a resolved run for this check. An exploratory spline refit, without
uncertainty intervals, puts run size plus the dependency features below chance on three of the
\NObsCorpora{} held-out corpora (\appref{app:d}).

What these results support is narrower than a shared invariant. Under one learner and on these
\NObsCorpora{} corpora, the direction run size carries flips on two held-out corpora, while the
dependency features keep their sign on all \NObsCorpora{}. That contrast is a property of the fitted
model as much as of the corpora. Under two other learners, the dependency features alone also fall
below chance on some held-out corpora (\appref{app:d}). Within a corpus, run size is
stronger than the dependency features alone on the size-strong corpora, so dependency does not
dominate everywhere.

\section{When a Dependency Layer Restates Run Size}
\label{sec:misread}

\textbf{A dependency layer can score well by restating run size.} Under the full-history assumption,
each step used every earlier step. Edge count and depth then follow from the step count, so the
dependency features read off that layer restate run size. A sparse layer can restate it too
(\secref{subsec:degenerate}). On SWE-Gym, the reading-free chain and the
full-history layer read nothing about reliance, yet both beat the recovered dependency layer
(Tables~\ref{tab:ladder} and~\ref{tab:gate}).

\subsection{How a Layer Collapses to Run Size}
\label{subsec:degenerate}

\begin{figure}[t]
\centering
\includegraphics[width=\columnwidth]{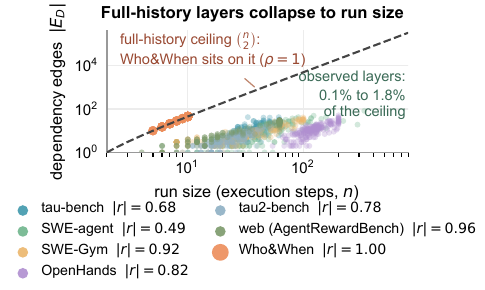}
\caption{\textbf{Inferred full-history dependencies collapse to run size; observed dependencies sit far below the ceiling.} Dependency-edge count $\lvert E_D\rvert$ versus run size $n$ per run, log-log, across the seven corpora; the dashed line is the full-history ceiling $\binom{n}{2}$. The $126$ failed Who\&When runs lie on the ceiling, every one at $\rho=1$, while the observed corpora sit roughly two to three orders below it (per-corpus medians $0.001$ to $0.018$)  yet still climb with $n$. A layer at the ceiling is run size in disguise; sitting below it does not establish the converse.}
\Description{Log-log scatter of dependency-edge count versus run size for seven agent corpora.
The Who and When corpus lies along the binomial-coefficient ceiling line, while the six
observed-dependency corpora form a cloud far below the ceiling that still rises with run size.}
\label{fig:degenerate}
\end{figure}

A full-history layer is a fixed function of run size. If each of a run's $n$ steps used every
earlier step, then
\[
  \begin{aligned}
    n_{\mathrm{dep}} \;=\; \lvert E_D \rvert &\;=\; \binom{n}{2} \;=\; \frac{n(n-1)}{2}, \\
    \mathrm{depth}(E_D) &\;=\; n - 1 .
  \end{aligned}
\]
A measured edge and an assumed one carry the same edge type (Figure~\ref{fig:twolayer}E), so the
graph alone cannot tell a real dependency from one the assumption invented. Observed layers track
run size too, though less tightly: their edge count correlates with step count at $\lvert r \rvert$
up to \MaxDepSizeCorr{}. They sit two to three orders of magnitude below the full-history ceiling
yet still climb with $n$ (Figure~\ref{fig:degenerate}, \appref{app:misread}). That is why
\secref{sec:signal} divides every feature by run length.

A sparse layer can restate run size as well. Define the saturation ratio as the edges a layer has
divided by the full-history edge count,
$
  \rho \;=\; \frac{n_{\mathrm{dep}}}{\binom{n}{2}} \in [0, 1] ,
$
which reaches $1$ for the full-history layer. The reading-free chain links each step to the step just
before it, so its ratio is only $2/n$. Yet each of its four features is $1/n$ or $1-1/n$, so across
runs they vary along one direction set by run size (\appref{app:within}). A small saturation ratio
therefore certifies nothing about whether a layer restates run size.

\subsection{Full-History and Counterfeit Layers in Transfer}
\label{subsec:gate}

\begin{table}[t]
\centering\footnotesize
\setlength{\tabcolsep}{4.pt}
\renewcommand{\arraystretch}{1.08}
\caption{\textbf{Swap the observed dependency layer for a full-history inferred one and the cross-corpus signal collapses.} The same size-normalized shape features on the observed layer (\textsc{aware}) and the full-history inferred layer (\textsc{+inf}, $\mathcal{A}_0$), against run size (\textsc{flat}); ROC-AUC, within corpus and pooled transfer. \textbf{Bold}: within-corpus set beats \textsc{flat}; \textit{italic}: below-chance transfer. The observed layer sits between $\rho=0.001$ and $\rho=0.018$, the inferred at $\rho=1.000$ by construction.}
\label{tab:gate}
\begin{tabular}{@{}l ccc c ccc@{}}
\toprule
 & \multicolumn{3}{c}{Within corpus} & & \multicolumn{3}{c}{Transfer (held-out)} \\
\cmidrule(lr){2-4}\cmidrule(lr){6-8}
Corpus & \textsc{flat} & \textsc{aware} & \textsc{+inf} & & \textsc{flat} & \textsc{aware} & \textsc{+inf} \\
\midrule
tau-bench  & 0.619 & \textbf{0.665} & 0.618          & & \textit{0.436} & 0.553 & \textit{0.443} \\
SWE-agent  & 0.628 & \textbf{0.713} & 0.595          & & 0.624          & 0.691 & 0.605 \\
SWE-Gym    & 0.663 & \textbf{0.804} & \textbf{0.828} & & \textit{0.359} & 0.590 & \textit{0.342} \\
OpenHands  & 0.698 & 0.695          & 0.698          & & 0.699          & 0.696 & 0.699 \\
tau2-bench & 0.722 & 0.706          & 0.716          & & 0.723          & 0.663 & 0.723 \\
web        & 0.765 & 0.756          & 0.754          & & 0.766          & 0.643 & \textit{0.287} \\
\bottomrule
\end{tabular}
\end{table}

\textbf{A full-history layer on the same steps carries only run size.} In each corpus we swap it in
for the recovered dependency layer, keeping the features, labels, splits and logistic model. Within
a corpus it then fails to improve on the run-size baseline on five of the \NObsCorpora{} corpora.
The exception is SWE-Gym, where the gain is run size in disguise. Given the four counts, this layer and the
reading-free chain each add only the reciprocal of the step count, so the two score alike there. In
transfer both fall below chance on three held-out corpora, one more than run size
(Table~\ref{tab:gate}, \appref{app:misread}).

\textbf{The saturation ratio flags such a layer but cannot clear one.} Across the corpora, the recovered layer holds a
small fraction of the full-history edges, and the full-history layer holds all of them
(Figure~\ref{fig:degenerate}). This makes the ratio a diagnostic to report beside the grade. Three
off-the-shelf graph networks match run size plus the dependency features within a corpus but trail
them in transfer (\appref{app:gin}). Like the logistic model, the edge-type-blind one is fooled by the
full-history layer on web.

\textbf{Two readings of the within-corpus gain remain.} Within a corpus, the preregistered test does
not separate the recovered layer from the degree-matched counterfeit (\secref{subsec:ladder}). That
counterfeit keeps how many earlier steps each step depends on but rewires which ones. Attachment is
therefore not a demonstrated source of the gain. On SWE-Gym, reliance evidence is not a demonstrated
source either: the full-history layer reads none and still scores higher. One reading credits the
gain to the degree sequence, each step's count of earlier sources, which the counterfeit keeps
(Figure~\ref{fig:twolayer}C). The other lets the recovered layer carry something the counts miss
when the corpus changes. Telling them apart takes the counterfeit scored in transfer.

\textbf{In transfer, the recovered layer leads the counterfeit on some held-out corpora.} It leads
on average, where within a corpus the mean margin was near zero. The recovered layer also stays
above chance on every held-out corpus, while the counterfeit falls below chance on SWE-Gym alone.
Since every layer is read together with run size, these counts cannot split the transferred signal
between the degree sequence and the attachment. The margin over the counterfeit measures only how
much transfer changes when the attachment is rewired, under one learner. In the transfer permutation
test, few or no degree-matched draws reach the recovered layer on SWE-Gym, tau-bench and web. Only
SWE-Gym separates after correction for multiple tests. Paired task resampling, registered after that
test and before any interval, separates all three after correction (\appref{app:n2transfer}). On
\NTwoTrInverts{} corpora the counterfeit scores higher.

\textbf{Withholding the task family reverses the largest transfer lead.} Preregistered
family-withheld transfer removes the held-out corpus's whole task family (coding; database and tool
use; web) from training. The recovered layer then shows no positive separation from the counterfeit
on any held-out corpus whose training pool changed. On SWE-Gym, where its lead was largest, the
recovered layer now trails the counterfeit, and under paired task resampling that reversal survives correction
for multiple tests (\appref{app:familytransfer}).

\section{Related Work}
\label{sec:related}

This paper meets two literatures that seldom cite each other, agent-graph systems and graph learning
on typed, temporal, and uncertain graphs; \appref{app:related} treats seven adjacent areas.
Orchestration and optimizable-agent graphs~\cite{zhuge2024gptswarm,hong2024metagpt,besta2024got}
represent a multi-agent system as a graph and tune it for task performance. None separates a typed
state-dependency layer from execution, grades an edge by how it is known, or measures a dependency
layer across corpora. A recent line trains a graph network on the workflow graph to predict
performance~\cite{zhang2025flora}. We reuse those workflow graphs in an exploratory comparison of
graph summaries against node and edge counts (\appref{app:declared}). The dependency layer alone is
a data-lineage or program-dependence graph~\cite{ferrante1987pdg,moreau2013provdm}. Network measures
of such graphs already predicted defects in software \cite{zimmermann2008network}, and agent
computation-graph views type them for optimization~\cite{acg2026survey,wang2025dyflow}. A lineage
graph answers what depended on what; the grade also answers how we know. Probing work corroborates
that the dependency object is recoverable~\cite{sun2026toolcall}: a low-capacity probe decodes
tool-call dependency structure from residual streams above both a random-label control task and a
positional baseline. The effect attenuates as call order alone becomes a sufficient proxy.

The two-layer representation also separates \grade{} from recent graph work on agent traces. CHIEF
builds a hierarchical causal graph and FALAT a typed step-dependency structure, both for failure
attribution on Who\&When; neither separates execution from a dependency layer graded by how each
edge is known \cite{wang2026chief,rafi2026falat}. Three later systems bound what is new here. EDGE distinguishes observational propagation edges from
intervention-validated edges and compares its validated graph against an edge-count-matched random
control, so an edge-evidence distinction already enters a control design \cite{hou2026edge}. Where
EDGE attributes errors over an error dependency graph, AIG attributes them over an adaptively built
influence graph and reports that richer trace representations improve attribution
\cite{bakish2026aig}. RUPA propagates trajectory-level uncertainty
over execution-event nodes on tau-2, Terminal-Bench-2 and GAIA, which overlaps our task
\cite{ma2026rupa}. None of
the three recovers a dependency layer over run steps in \NObsCorpora{} corpora, and none reports a
control that holds each step's degree fixed while varying the attachment. \emph{From Agent Traces to Trust} frames agent execution as a typed graph for evidence tracing, but
stays a tracing and lineage survey without a graded two-layer model \cite{wang2026tracestotrust}.
AgentArmor maps traces into CFG, DFG and PDG for prompt-injection defense, a security analysis layer
with no graded dependency layer tied to run-failure prediction \cite{wang2025agentarmor}.

\section{Conclusion}
\label{sec:conclusion}

A trace records what an agent did; \grade{} adds a graded record of what each step relied on and
measures its gain over run size. Over a linear run-size baseline, the recovered dependency layer
improves failure prediction on three of the six corpora, but the gain depends on the learner and task sampling.
On held-out corpora it stays above chance where run size does not, under one learner and with the
same task-sampling caveat. Graph structure can also pass for signal it does not carry: a layer
inferred under full history restates run size. Within a corpus, a preregistered test does not
separate the recovered layer from a degree-matched counterfeit. The counterfeit keeps how many
earlier steps each step depends on but rewires which ones. In transfer the recovered layer leads it
on some held-out corpora and trails it on others, and its largest lead reverses when the held-out
corpus's task family is withheld. A gain over run size does not by itself show that the attachment
carries it. Reuse the protocol: grade each edge, keep a run-size baseline, and score degree-matched
attachment controls within a corpus and in transfer.

\clearpage

\section*{Limitations}

\textbf{What the dependency layer adds over run size is fragile.} Over a linear run-size baseline it
adds failure-prediction signal on three of \NObsCorpora{} corpora. That gain depends on the learner
and on task sampling. Resampling tasks can leave its sign unresolved, and a corpus that never
repeats a task cannot test repeated-task leakage. Against the composition baseline, which also
counts reads, writes and resources, the gain shrinks under the logistic model with run-level folds.
Its point estimates there turn negative under flexible learners.

\textbf{Transfer holds under one learner only.} On every held-out corpus the dependency features
alone stay above chance, while run size falls below chance on two. No prediction was preregistered
for this pattern. An exploratory spline refit carries no uncertainty intervals. Under it, the
dependency features alone fall below chance on four of \NObsCorpora{} held-out corpora, and run size
plus the dependency features on three. No transfer check rescales run size per corpus or builds the
dependency-only block from a counterfeit layer, so part of the contrast may reflect feature scaling.

\textbf{Counterfeit layers give the recovered attachment no stable credit.} Within a corpus, the
preregistered test does not separate the recovered dependency layer from the degree-matched
counterfeit. That counterfeit keeps how many earlier steps each step depends on but rewires which
ones, and two of the four features cannot move under it. In transfer, the recovered layer separates
from it on one to \NTwoTrSeparates{} held-out corpora, depending on the test, and the counterfeit
scores higher on \NTwoTrInverts{}. Family-withheld transfer leaves no positive separation on any
held-out corpus whose training pool changes, and the recovered layer's largest lead turns negative.
These results do not show that the attachment or the grade is inert, and they license no claim beyond the evaluated
extraction, learner, corpora and counterfeits.

\textbf{The corpora are narrow and the runs are short.} Coverage leans on coding and database tasks.
Three corpora mix generating models, and no analysis separates a gain from model identity. Runs
are short and observed reliance is mostly local; long-range reliance, where run size may become a
stronger baseline, is untested.

\textbf{The grading rules and graph models are preliminary.} Observed-grade edges rest on heuristic
attachment rules, since no agent corpus has ground-truth reliance labels at scale. The per-edge
grade is a proposed schema: no experiment mixes grades within a graph, so whether a model can use it
is untested. We compare only off-the-shelf graph networks, and which architectural change would
close their transfer deficit against run size plus the dependency features is open. \appref{app:limitsdetail}
states each limitation in full.

\section*{Ethical Considerations}

This work studies a graph representation of LLM agent runs, evaluated on public benchmarks, and it
introduces no new data collection, human subjects, or personally identifying information. By predicting failures and
localizing faults, the representation is meant to make deployed agents more reliable and auditable.
Beyond the general dual-use nature of any reliability tooling, we identify no specific risk: the
structural signals it reads come from a run's own trace, already available to whoever operates the
agent. Where the declared grade adds instrumentation, the dependency logs it records (file, database, and tool I/O) can carry sensitive payloads from the underlying task, so a deployment should minimize and access-control them.

\bibliography{references,yue-zhao,working}

\clearpage
\appendix

\section*{Supplementary Material for \grade{}}
\setcounter{figure}{0}
\setcounter{table}{0}
\renewcommand{\thefigure}{A\arabic{figure}}
\renewcommand{\thetable}{A\arabic{table}}
\setcounter{footnote}{0}     

\section{Formal Class and Correspondence}
\label{app:formal}

\subsection{The Formal Tuple and Recovery Maps}

A run is a directed, typed, temporal multigraph
\[
  G \;=\; (V,\ E_X,\ E_D,\ \tau,\ t,\ \sigma),
\]
with a typing map, a time map, and a source map
\[
  \begin{aligned}
    \tau &\colon V \to \{\textsf{agent}, \textsf{decision}, \textsf{tool}, \textsf{resource}\}, \\
    t &\colon V \to \mathbb{R}_{\ge 0}, \\
    \sigma &\colon E_D \to \{\textsf{observed}, \textsf{declared}, \textsf{inferred}\}.
  \end{aligned}
\]
The map $\tau$ labels each node and $t$ orders nodes in time, with resource nodes versioning the
state they hold. The execution layer $E_X \subseteq V \times V$ is labeled by
$\kappa\colon E_X \to \{\textsf{emit}, \textsf{handoff}\}$: an agent emits its decision and tool
nodes, and control hands off between consecutive steps. The dependency layer
$E_D \subseteq V \times V$ holds reliance edges,
\[
  (u,v) \in E_D \;\Rightarrow\; t(u) < t(v),
\]
meaning step $v$ used state produced or held by the earlier node $u$. The source map $\sigma$
records, per edge, which grade of the attachment model it carries.

The two layers arise from a trace $T$ in different ways. The execution layer is a deterministic
read-off,
\[
  E_X = f_{\mathrm{exec}}(T).
\]
The dependency layer is not fixed by $T$ alone. Where read and write events $R(T)$ are available,
whether already present in the trace content or emitted by added instrumentation, the dependency
edges are recovered from them,
\[
  f_{\mathrm{rec}}(R(T)) \subseteq E_D,
\]
and $\sigma$ marks each such edge \textsf{observed} or \textsf{declared} accordingly. Where no
access events exist, the layer is inferred under an assumption $\mathcal{A}$,
\[
  E_D = f_{\mathrm{infer}}(T, \mathcal{A}), \qquad \sigma \equiv \textsf{inferred},
\]
the weakest assumption $\mathcal{A}_0$ being full history:
\[
  (u,v) \in E_D \quad \text{for every } u \text{ with } t(u) < t(v).
\]

\subsection{Related Views as Fragments of One Graph}

Execution-side fragments are the common case. A reason-act trace~\cite{yao2023react} is
the execution chain of a single agent expanded over its tool calls,
and an orchestration or optimizable-agent graph~\cite{zhuge2024gptswarm,hong2024metagpt} is the
agent-level execution graph used to coordinate agents or to optimize task performance. Some of
these carry dependency-like edges. GPTSwarm's single edge set is at once execution topology and
information flow, and that paper states the gap the two layers keep apart, that ``many routines
might be designed to ignore the input and operate solely in the context provided by the predecessor
nodes.'' A computation graph~\cite{acg2026survey} carries one undifferentiated edge set in its
definition, with the control, data and communication split described in prose and no typing function
that separates them. In each of these the dependency layer stays partial and carries no per-edge
grade. The mirror-image fragment drops the execution layer instead: a data-lineage or PROV graph
keeps what was read and written and carries no per-edge grade, and its activity ordering is
entity-mediated in both directions: one inference derives that ordering from generation composed
with use, and the converse makes a directly asserted ordering imply an entity that one activity
generated and the other used~\cite{cheney2013provconstraints}. The Open Provenance Model calls the
same composition's summary lossy~\cite{moreau2011opm}. So two steps that ran in sequence and
exchanged nothing get no edge at all. Layered lineage-capture systems already separate what the
system observes from what an application discloses through an
interface~\cite{muniswamyreddy2009layering}; the grade carries that distinction onto agent dependency
edges and adds a rung for edges posited under an assumption. Agent-graph anomaly
detectors~\cite{he2025sentinelagent,zhou2025guardian} build an execution graph for a security
objective and likewise leave state dependency partial; GUARDIAN is homogeneous, with one node kind
and one edge kind, and SentinelAgent pairs a statically declared graph with a dynamically observed
one.

\section{The Two-Layer Object}
\label{app:object}

Neither layer is the contribution by itself. The execution layer alone is ordinary control flow,
the directed record of which step followed which that any log or call graph already supplies;
representing it is established, and \secref{sec:representation} reads prior agent-graph work
as projections onto it. A dependency layer on its own is a data-lineage graph, a
reads-and-writes relation over evolving state of the kind PROV-style and dataflow systems have
recorded for years. The object this paper adds is the pair, joined by a per-edge observability
grade: every dependency edge is stamped observed, declared, or inferred, so the model knows how
each reliance edge was obtained and can price the dependency layer against the execution one. The
attachment model (\secref{sec:representation}) is that stamp, and the marginal-lift estimand
is the price.

Two analyses run on the joined object, one per layer, and each is scored on its own task. The
dependency layer is scored on run-level failure prediction, reported as marginal lift over run size
(\secref{sec:signal}). A gold final state scores the two database corpora, issue resolution the
three software-engineering corpora, and task outcome the web corpus. On Who\&When the execution
layer is scored on a different task. A score learned over eight execution-layer features ranks a
failed run's labeled faulting step against a position prior (\appref{app:loc}). Within a
single edge set the observability grade keeps an assumed edge distinct from a logged one, so a
method reading the graph can treat them differently. Neither analysis sorts runs into coordination
and reliance categories, so Table~\ref{tab:twomodes} states the design rationale for carrying two
layers and reports no measured mapping from failure type to layer. Whether the pair is required
also goes untested. Marginal lift prices one layer against a summary of the other, and no
experiment here asks whether both must sit on one node set for either result to hold. Attributing a
run to its failure mode awaits a corpus with observed dependencies and step-level fault labels, a
gap the Limitations section already records.

\begin{table*}[t]
\centering\footnotesize
\setlength{\tabcolsep}{6pt}
\renewcommand{\arraystretch}{1.15}
\caption{\textbf{The two edge layers and the analysis each one was evaluated on.} Failure prediction over \NObsCorpora{} corpora evaluates the dependency layer, and ranking a failed Who\&When run's labeled faulting step evaluates the execution layer. The failure-mode row states the design rationale for carrying two layers; no experiment here assigns a run to a failure mode.}
\label{tab:twomodes}
\begin{tabular}{@{}>{\raggedright\arraybackslash}p{0.13\textwidth} >{\raggedright\arraybackslash}p{0.39\textwidth} >{\raggedright\arraybackslash}p{0.39\textwidth}@{}}
\toprule
 & \textbf{Dependency layer} \textit{(the costly signal)} & \textbf{Execution layer} \textit{(the free base)} \\
\midrule
Records & State reliance: what each step read or wrote & Control flow: which step ran, in what order \\
Failure mode & \emph{Reliance}: the state a step used has gone bad & \emph{Coordination}: control moves wrongly \\
Surfaces as & Size-normalized dependency shape predicts failure & Execution topology ranks the labeled faulting step \\
Tested in & \secref{sec:dependency}, six corpora & \appref{app:loc}, Who\&When \\
\bottomrule
\end{tabular}
\end{table*}

\subsection{Keeping the Marginal-Lift Estimand Honest}

One choice keeps the estimand honest and one precaution is weaker than it looks. The comparison
is nested, flat against flat-plus-dependency on identical cross-validation splits, so
$\Delta_{\mathrm{dep}}$ isolates what the dependency block adds to the counts. The features are
also divided by run length, which rescales them but does not free them of run length: under the
full-history assumption the normalized edge count is $(n-1)/2$, still linear in $n$. So
normalization does not establish that a gain reflects information beyond run size, and the paper
does not rest on it: on OpenHands and the web corpus a dependency-only model reaches essentially
the flat AUC and adding the block moves the mean down, which is what redundancy looks like under
this probe. The estimand is allowed to
come out near zero. On the three size-strong corpora $\Delta_{\mathrm{dep}}$ lands at or just below
zero (\secref{sec:signal}). On the other three the lift is positive, and what it clears is a free run-size
summary drawn from the execution layer rather than chance; it does not clear every function of run size
(\secref{sec:signal}). The permutation diagnostic of \appref{app:ordering} finds the link between
baseline strength and gain compatible with arithmetic coupling alone.

\section{Within-Corpus Protocol and Ablations}
\label{app:within}

\subsection{Corpus Cards}
\label{app:corpuscard}

Table~\ref{tab:corpuscard} lists, for each corpus, the public release and revision, the models that
generated the runs, the rule that selected them, and the share of runs that carry no dependency edge.
The anonymized supplementary artifact holds the loaders that apply these rules and a machine-readable
card per corpus.

The observed grade records where the evidence for an edge
came from, and three different rules realize it here. On tau-bench a consequential write attaches to
\emph{every} prior database read, a conservative upper bound rather than a keyed match, because the
trace does not label which read a write relied on. In the file and web corpora an edge links each
access to the most recent earlier access to the same resource, which is resource-keyed. On tau2-bench
the rule is identifier-keyed: a step attaches to the most recent earlier step that produced an
identifier it consumes. Attachment rules affect density; each graph records its rule, which results
should name.

\begin{table*}[t]
\centering\footnotesize
\setlength{\tabcolsep}{2.5pt}
\caption{Corpus cards. Every corpus keeps runs with at least two steps; label $1$ is a failed run.
Only the tau-bench loader pins its revision; for the others the table gives the revision the rebuild
resolved. The web cohort was drawn from a copy of the release that lacks seven trajectory files, which
the card lists. Task counts use corrected identifiers; on the archived web cohort,
Figure~\ref{fig:within}(b) counts $202$. The SWE-agent, SWE-Gym and OpenHands scaffolds are read from
the release names; their machine-readable cards leave the scaffold field as not recorded. Where a release states no license,
its benchmark code is MIT (tau-bench, tau2-bench) or Apache 2.0 (SWE-Gym).}
\label{tab:corpuscard}
\begin{tabular}{@{}>{\raggedright\arraybackslash}p{1.75cm}*{6}{>{\raggedright\arraybackslash}p{2.13cm}}@{}}
\toprule
 & tau-bench & SWE-agent & SWE-Gym & OpenHands & tau2-bench & web \\
\midrule
Release & AgentSuite/ tau-bench-trajectories & nebius/ SWE-agent-trajectories & SWE-Gym/ OpenHands-Sampled-Trajectories & nebius/ SWE-rebench-openhands-trajectories & AgentSuite/ tau2-bench-trajectories & McGill-NLP/ agent-reward-bench \\
Revision & \texttt{382e57d} & \texttt{68195a1} & \texttt{baf3a4e} & \texttt{3545538} & \texttt{07264f1} & \texttt{b6d17e6} \\
License & none stated & CC BY 4.0 & none stated & CC BY 4.0 & none stated & research terms of use \\
\addlinespace[2pt]
Models (runs) & Claude 4.5 Sonnet (thinking), GPT-4.1, GPT-4o-mini, Kimi-K2-Instruct ($165$ each) & swe-agent-llama-70b ($570$), swe-agent-llama-8b ($30$) & GPT-4o (2024-08-06), one sampled run ($376$) & Qwen3-Coder-480B ($600$) & Qwen3-235B-A22B-Thinking ($278$) & Claude 3.7 Sonnet ($200$), Qwen2.5-VL-72B ($199$), GPT-4o ($103$), Llama-3.3-70B ($98$) \\
\addlinespace[2pt]
Scaffold & not recorded & SWE-agent & OpenHands & OpenHands & not recorded & GenericAgent \\
\addlinespace[2pt]
Selection & all runs of four model files with a database check & first $600$ runs of shard $0$ & largest balanced run cohort, $188$ per class & first $600$ runs of shard $0$ & largest balanced cohort of the first ten model files, $139$ per class & first $600$ annotated runs, WebArena and VisualWebArena \\
\addlinespace[2pt]
Runs / tasks & $660$ / $165$ & $600$ / $35$ & $376$ / $376$ & $600$ / $574$ & $278$ / $278$ & $600$ / $203$ \\
Failed & $363$ ($55.0\%$) & $540$ ($90.0\%$) & $188$ ($50.0\%$) & $312$ ($52.0\%$) & $139$ ($50.0\%$) & $399$ ($66.5\%$) \\
Steps, median [IQR] & $29$ [$25$, $38$] & $34$ [$22$, $56.5$] & $28$ [$14.75$, $50$] & $122$ [$101.5$, $150.5$] & $22$ [$16$, $29$] & $26$ [$12$, $62$] \\
No dependency edge & $119$ ($18.0\%$) & $74$ ($12.3\%$) & $3$ ($0.8\%$) & $0$ ($0.0\%$) & $23$ ($8.3\%$) & $25$ ($4.2\%$) \\
\bottomrule
\end{tabular}
\end{table*}

\subsection{Probe Design}

We predict run failure with standardized logistic regression scored by ROC-AUC,
the area under the receiver operating characteristic curve. The quantity of
interest is the marginal-lift estimand from \secref{sec:twomodes}: the
increase in AUC when the dependency block joins a model that already holds the free
execution-layer size features. The baseline model, \textsc{flat}, uses run size
alone, the counts of steps, tool calls, decisions, and agents, which is exactly
what the observed execution layer hands over at no cost. Within each corpus the agent count is
constant and decisions equal steps minus tool calls; on SWE-agent, OpenHands and web, tool calls
also equal decisions. The standardized \textsc{flat} block therefore has rank two on tau-bench,
SWE-Gym and tau2-bench, and rank one, run length alone, on the other three corpora. The treatment model,
\textsc{flat}+dep, adds the four size-normalized dependency features defined below. We
divide those
features by run length, which removes the raw growth of dependency counts with run
length without making them independent of it. Under the full-history assumption the
collinearity reaches an extreme and the layer becomes a deterministic function of run
size, the normalized edge count standing at $(n-1)/2$; that degenerate regime is
analyzed in \secref{sec:misread}. With observed dependencies the rescaling is a
precaution rather than a separation, so the nested comparison, and not the
normalization, is what makes $\Delta_{\mathrm{dep}}$ interpretable.

Write $s_1,\dots,s_n$ for the run's execution steps in temporal order, and
$D = \{(i,j) : (s_i,s_j) \in E_D\}$ for the dependency edges among them. Edges run
earlier to later as in \appref{app:formal}, so $(i,j) \in D$ implies $i < j$ and records that
step $j$ accessed a resource or identifier that step $i$ produced or accessed earlier, under the
corpus's attachment rule. The four features are
\[
  \begin{aligned}
    \phi_{\mathrm{dens}} &\;=\; \lvert D \rvert / n,
      \quad
    \phi_{\mathrm{depth}} \;=\; \mathrm{depth}(D) / n, \\
    \phi_{\mathrm{blast}} &\;=\; \tfrac{1}{n} \max_{i} \bigl\lvert \{\, j : (i,j) \in D \,\} \bigr\rvert, \\
    \phi_{\mathrm{audit}} &\;=\; \tfrac{1}{n} \max_{j} \bigl\lvert \{\, i : (i,j) \in D \,\} \bigr\rvert,
  \end{aligned}
\]
where $\mathrm{depth}(D)$ counts the edges on the longest directed path. A large
$\phi_{\mathrm{blast}}$ says that one step is relied on by a large share of the run, so a
fault in it can reach many later steps. Where $\phi_{\mathrm{audit}}$ is large, one step
draws on a large share of the run, so checking that step means reading most of the
history. The reference implementation calls the four \texttt{dep\_edges\_per\_step},
\texttt{rel\_dep\_depth}, \texttt{max\_blast\_share} and \texttt{max\_audit\_share}, in
that column order. It stores each edge reversed, from the later step to the earlier one,
so the blast count is an in-degree there and the audit count an out-degree.

Two cases make the arguments of \secref{sec:misread} checkable in one line. Under
$\mathcal{A}_0$ every earlier step is an endpoint, so $\phi_{\mathrm{dens}} = (n-1)/2$ and
$\phi_{\mathrm{depth}} = \phi_{\mathrm{blast}} = \phi_{\mathrm{audit}} = (n-1)/n$, all
four functions of $n$ alone. For the sparse chain, where each step attaches only to its
predecessor, $\phi_{\mathrm{dens}} = \phi_{\mathrm{depth}} = (n-1)/n$ and
$\phi_{\mathrm{blast}} = \phi_{\mathrm{audit}} = 1/n$ for $n \ge 2$. Each run's row is
then $\bigl(1 - \tfrac{1}{n},\; 1 - \tfrac{1}{n},\; \tfrac{1}{n},\; \tfrac{1}{n}\bigr)$,
so across runs of differing length the raw block has rank $2$. Centering the columns
leaves every row a multiple of $(-1,-1,1,1)$, so the centered rank is $1$. The
full-history block keeps rank $2$ under the same centering, over runs of three or more
distinct lengths. Its first column is affine in $n$ while the other three are affine in
$1/n$.

We evaluate every model under five-fold stratified cross-validation repeated over five
seeds, and report seed-block 95\% confidence intervals across the five seeds. A gain
whose interval excludes zero clears zero across split seeds; the interval measures split sensitivity
on a fixed corpus, not the sampling of tasks. On the within-corpus AUCs of Table~\ref{tab:ladder}, in that table's corpus order, the
seed-block $95\%$ half-widths are $\pm0.003$, $\pm0.003$, $\pm0.007$, $\pm0.003$, $\pm0.005$ and
$\pm0.002$ for \textsc{flat}, and $\pm0.006$, $\pm0.009$, $\pm0.005$, $\pm0.007$, $\pm0.002$ and
$\pm0.007$ for \textsc{aware}.

\begin{figure}[t]
\centering
\captionsetup{skip=4.5pt}
\includegraphics[width=\columnwidth]{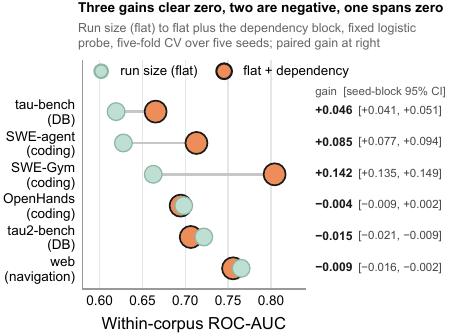}
\Description{A horizontal dumbbell plot, one row per corpus, from the run-size baseline AUC to the
size-plus-dependency AUC under the same logistic probe, with the difference and its seed-block
interval printed in a right-hand column.}
\caption{Within-corpus ROC-AUC under the fixed logistic probe, from \textsc{flat} to \textsc{flat}
plus the size-normalized dependency block, in Table~\ref{tab:ladder}'s corpus order. Order and gain share the
\textsc{flat} baseline, so they are arithmetically coupled (permutation $p=0.56$, \appref{app:ordering}).}
\label{fig:keystone}
\end{figure}

\subsection{Dependency Shape Adds Beyond Revisit Density}

Granting the within-corpus lift of \secref{sec:signal} (Figure~\ref{fig:keystone}), a skeptic can still doubt its content. Perhaps the
dependency block only counts how often a run revisits state, so failed runs touch
the same file or row more often and the gain is revisit volume rather than
dependency structure. We separate volume from structure. The dependency block
splits into one revisit-density feature, dependency edges per step, and three shape
features: relative reliance-chain depth, the largest blast hub share (the
most-relied-upon node's share of downstream steps), and the largest audit hub share
(the step that draws on the most prior state). We then ask whether shape adds
anything once density is in the model. It does, on all three corpora where
dependency helps: \TauShapeGain{} on tau-bench, \SweShapeGain{} on SWE-agent, and
\SweGymShapeGain{} on SWE-Gym, five of five seeds with the seed-block 95\% interval
excluding zero.

The three corpora reach that result by different routes, and the difference bounds
the claim. tau-bench is the clean case: adding revisit density to run size leaves the
AUC where it was, \TauFlat{} to \TauFlatDens{}, a paired gain whose seed-block interval
$[-0.002, +0.003]$ spans zero, and the lift appears only when chain depth and hub
concentration enter (\TauFlatDep{}). There the gain is shape rather than revisit
count. On SWE-agent and SWE-Gym the density feature itself already raises AUC above
the flat baseline, and shape adds a further \SweShapeGain{} and \SweGymShapeGain{}
on top. There revisit density is part of the signal and shape is the part it
misses. The honest summary is therefore narrower than a blanket claim that the gain
ignores revisit count: dependency shape carries information beyond revisit density
on every corpus where the dependency layer helps, while revisit density itself
contributes on two of the three.

\subsection{Dependency Alone Matches Run Size Where Size Is Strong}

The redundancy reading of the three size-strong corpora makes a testable claim. If
the dependency layer is redundant with size there, then dependency features on their
own, with the size counts removed, should still predict failure. They do. A dependency-only
model reaches 0.694 on OpenHands against the flat \OhFlat{} from size and 0.755 on the web
corpus against the flat \WebFlat{}, matching size on both; on tau2-bench it is weaker at 0.645
against the flat \TauTwoFlat{} but stays well above chance.
The dependency layer still predicts failure by itself; it is simply redundant with
size, which is why it adds no lift once size is in the model. Redundancy here
is overlap between two informative blocks.

The same test draws out the complementary case. On tau-bench the dependency-only
model (0.569) is weaker than size alone (\TauFlat{}), yet the combined model clears
both at 0.665: dependency and size each catch failures the other misses, so they add
instead of overlap. The split between tau-bench, where the two blocks complement,
and OpenHands and the web corpus, where they overlap, is the split the marginal-lift
estimand exists to read. It is also why we price the costly dependency layer against
the free execution layer rather than reporting either block alone.

\subsection{Matched Learner Controls for Run Size}
\label{app:c1}

\begin{figure*}[t]
\centering
\includegraphics[width=\textwidth]{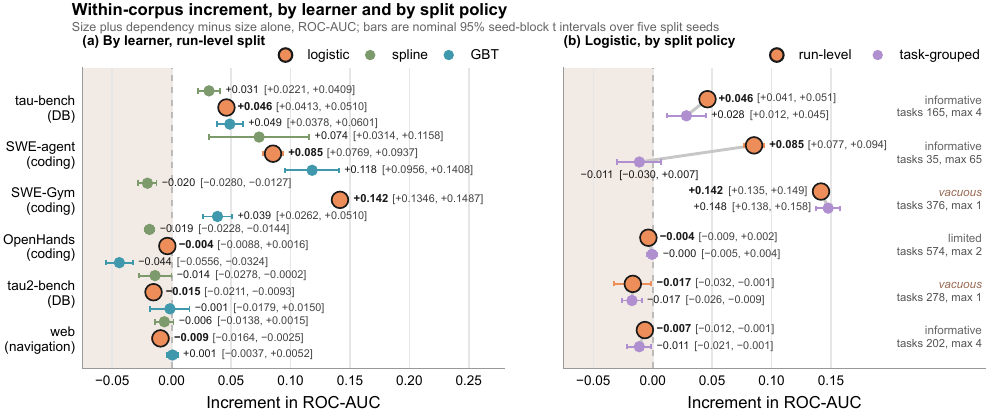}
\Description{Two panels of horizontal interval plots, one row per corpus. Panel (a) shows the within-corpus increment of size plus dependency over size alone under three learners with seed-block intervals; panel (b) compares the run-level split with task-grouped folds for the logistic probe and marks each corpus informative, limited, or vacuous.}
\caption{Within-corpus matched increments in ROC-AUC, size plus dependency minus size alone. (a) Under three learners; bars give nominal 95\% seed-block $t$ intervals for variation across the five split seeds. (b) Under the logistic probe, with a run-level split and with folds grouped by task. \emph{Tasks} counts distinct task identifiers and \emph{max} the largest number of runs any one task contributes. A corpus whose tasks each carry one run already holds out whole tasks under either policy, so the comparison between its two markers tests nothing about repeated-task leakage and is marked vacuous rather than counted as a check that passed. In (b), the tau2-bench and web rows run on historical inputs, described in the text, which is why their run-level markers differ from the logistic markers in (a). Both rows are retained as historical checks.}
\label{fig:within}
\end{figure*}

Figure~\ref{fig:within}(a) compares size plus dependency features with size alone using the same learner
and matched splits. The size block contains the four recorded counts: steps, tool calls, decisions, and agents. The augmented arm adds the recorded dependency feature block. All results use the corrected dependency cache and the shared web cohort used by the primary analysis.

Each arm uses five seeds and five stratified run-level folds per seed. We average the five fold AUCs within each seed and form paired differences $d_s$. The reported interval is $\bar d \pm 2.776\,\mathrm{sd}(d_s)/\sqrt{5}$. These nominal intervals describe split sensitivity on the fixed corpus; repeated folds reuse runs, and the intervals do not quantify sampling of new tasks or corpora. No adjustment across the 18 comparisons is applied. Grouped evaluation is a separate analysis.

Logistic regression uses training-only standardization, balanced classes, liblinear, and $C=1$. The spline model applies training-only variance filtering, cubic splines with five quantile knots per retained input, linear extrapolation, and standardization before the same logistic learner. The augmented spline arm also expands the dependency inputs. GBT uses 200 iterations, learning rate $0.05$, at most 15 leaves, minimum leaf size 20, L2 regularization 1, balanced classes, and no early stopping. The learner configurations are fixed; this comparison does not search all nonlinear size models.

SWE-Gym has a positive logistic increment ($+0.142$), a negative spline increment ($-0.020$), and a positive GBT increment ($+0.039$), with all three intervals excluding zero. Mixed point-estimate signs occur on two of six corpora, SWE-Gym and web. Both flexible-learner intervals on web include zero, leaving its reversal unresolved. OpenHands and tau2-bench have negative point estimates under all three learners. On OpenHands, the spline and GBT intervals exclude zero; on tau2-bench, the logistic and spline intervals exclude zero. A flexible-learner increment has an interval entirely above zero on tau-bench, SWE-agent, and SWE-Gym. These comparisons do not identify dependency information independent of every function of run size.

For transparency, two selected cross-learner contrasts on SWE-Gym are also retained as exploratory checks. Selecting the strongest dependency arm (GBT) and strongest size arm (spline) gives $+0.012$ AUC; comparing the logistic dependency arm with the selected size spline gives $-0.028$. Both choices use the observed evaluation results, change the learner across arms, and lack a selection-adjusted interval. The size spline reaches $0.832$ AUC, compared with $0.804$ for the logistic dependency arm; that comparison cannot replace a matched increment.

\textbf{An execution-side composition baseline.} A stronger baseline adds to the four counts the
numbers of reads, writes and distinct resources, a repeat count, per-tool call counts, $\log n$ and
$1/n$. Under the logistic probe the increment over it falls from $+0.046$ to $+0.018$ on tau-bench
(task-grouped, from $+0.028$ to $+0.014$, interval $[+0.002, +0.026]$), from $+0.085$ to $+0.065$ on
SWE-agent (task-grouped, from $-0.011$ to $+0.010$, interval spanning zero) and from $+0.142$ to
$+0.012$ on SWE-Gym, and it stays at or below zero on the other three corpora. Under the spline and the
boosted tree with run-level folds, each of the three point estimates is negative and no interval lies
above zero. With task-grouped folds the spline gives $-0.011$ ($[-0.019, -0.003]$) on tau-bench and
$+0.022$ ($[+0.013, +0.030]$) on SWE-agent, where both spline arms score below chance ($0.439$ and
$0.461$); the boosted tree gives $+0.001$ and $-0.009$, both intervals spanning zero. This
comparison and the task-resampled intervals of \appref{app:c2} ran under a protocol written before
their values were computed. Its reproduction gate stops an analysis on failure, and it stopped both
on tau2-bench, where runs with identical features and opposite labels receive exactly tied
probabilities in the rerun. The gate was then relaxed to accept such ties. Every printed value it
checks is unchanged, and the tau2-bench values of both analyses use the rerun's resolution of the
ties. Before that relaxation, scratch runs had computed this comparison's values on the other five
corpora, which pass the original gate, so the relaxation cannot move them. Earlier code-path checks
of the task-resampled intervals, with few replicates, placeholder web identifiers and, in one case,
a loosened gate, were never reported or used. The supplementary
deviation log records the relaxation, the scratch runs and the code-path checks, and states that the
relaxation was decided on the gate evidence alone.

\subsection{Flexible Size Baselines}
\label{app:flexsize}

The size block of \appref{app:c1} enters the probe linearly. Table~\ref{tab:flexsize} adds two
flexible size baselines, fixed with their reporting rule before any value was computed: B1 adds $1/n$
and $\log n$ to the four counts, and B2 adds a cubic spline basis in $n$. Each treatment adds the four
dependency features to its baseline, with the learner, folds and seed-block intervals of
\appref{app:c1}. Task-grouped folds follow \appref{app:c2} on the corrected task identifiers, so web's
task-grouped B0 row is the current-cache value that \appref{app:c2} reports, not the archived row of
Figure~\ref{fig:within}(b). The rule calls an increment separated from these functions of run size
only when its B1 interval lies above zero under run-level folds and under the corpus's grouping
policy: task-grouped folds where tasks repeat, and folds grouped by repository on SWE-Gym. Only
tau-bench meets it. Under B1 and B2 alike, its seed-block intervals lie above zero under both fold
policies, and its task-level intervals include zero. With run-level folds, its B0 and B1 task-level
intervals (Tables~\ref{tab:within} and~\ref{tab:flexsize}) agree at three decimals and differ at four:
$[-0.0053, +0.0996]$ against $[-0.0048, +0.1000]$. SWE-Gym's increment turns negative under both
baselines. The composition baseline of \appref{app:c1} contains B1's terms, yet over it SWE-Gym's
increment is $+0.012$; that baseline alone scores $0.798$ against B1's $0.827$, so the sign depends on
which other counts share the model. SWE-agent's B1 increment stays positive under run-level folds and
loses its interval under task grouping; its task-grouped B2 interval clears zero by $0.0003$, between
two arms below chance (AUC $0.433$ and $0.446$). Folds grouped by repository cannot be formed on
SWE-Gym: only two of its $11$ repositories hold a failed run, so three of every five test folds hold
resolved runs only. That policy yields no AUC, no claim rests on it, and SWE-Gym already fails the
run-level condition.

\begin{table*}[t]
\centering\footnotesize
\setlength{\tabcolsep}{4pt}
\caption{Within-corpus increment of the dependency block over three size baselines under the logistic
probe: B0 is the four counts, B1 adds $1/n$ and $\log n$, and B2 adds a cubic spline basis in $n$ with
five knots at training-fold quantiles. Brackets are seed-block 95\% intervals; the last column
resamples tasks for the pooled out-of-fold B1 increment. \emph{run} and \emph{task} are run-level and
task-grouped folds. A dash marks a corpus with too few task clusters for that interval.}
\label{tab:flexsize}
\begin{tabular}{@{}llcccc@{}}
\toprule
Corpus & Folds & B0 & B1 [95\%] & B2 [95\%] & Task CI, B1 \\
\midrule
tau-bench & run & $+0.046$ & $+0.046$ $[+0.041, +0.052]$ & $+0.047$ $[+0.037, +0.057]$ & $[-0.005, +0.100]$ \\
 & task & $+0.028$ & $+0.028$ $[+0.009, +0.047]$ & $+0.030$ $[+0.014, +0.047]$ & $[-0.022, +0.084]$ \\
\addlinespace[2pt]
SWE-agent & run & $+0.085$ & $+0.125$ $[+0.110, +0.141]$ & $+0.097$ $[+0.074, +0.120]$ & --- \\
 & task & $-0.011$ & $+0.006$ $[-0.010, +0.021]$ & $+0.013$ $[+0.000, +0.026]$ & --- \\
\addlinespace[2pt]
SWE-Gym & run & $+0.142$ & $-0.006$ $[-0.011, -0.002]$ & $-0.006$ $[-0.009, -0.002]$ & $[-0.024, +0.021]$ \\
\addlinespace[2pt]
OpenHands & run & $-0.004$ & $-0.004$ $[-0.009, +0.001]$ & $-0.003$ $[-0.008, +0.001]$ & $[-0.013, +0.005]$ \\
 & task & $-0.000$ & $-0.000$ $[-0.005, +0.004]$ & $-0.000$ $[-0.001, +0.001]$ & $[-0.009, +0.009]$ \\
\addlinespace[2pt]
tau2-bench & run & $-0.015$ & $-0.018$ $[-0.024, -0.013]$ & $-0.021$ $[-0.036, -0.007]$ & $[-0.034, -0.003]$ \\
\addlinespace[2pt]
web & run & $-0.009$ & $-0.003$ $[-0.011, +0.005]$ & $-0.008$ $[-0.012, -0.004]$ & $[-0.015, +0.013]$ \\
 & task & $-0.014$ & $-0.008$ $[-0.015, -0.000]$ & $-0.016$ $[-0.024, -0.009]$ & $[-0.018, +0.007]$ \\
\bottomrule
\end{tabular}
\end{table*}

\subsection{Grouping the Folds by Task}
\label{app:c2}

A run-level split can place two runs of the same task on both sides of a fold, so a model may be
scoring task identity rather than run structure. We compare run-level folds against
\textsc{StratifiedGroupKFold} on the saved task identifier, five folds and seeds $0$ to $4$, with
the size and size-plus-dependency arms sharing folds within each policy and the learner and feature
blocks of \appref{app:within} unchanged. This comparison runs on the archived C2 cache. Four
corpora match the cache used elsewhere in this paper: tau-bench, SWE-agent, SWE-Gym and OpenHands.
Two do not. The tau2-bench row carries historical dependency features that the producer fix
superseded. The web row runs on an archived $600$-run cohort that shares $599$ runs with the
current one, and those shared runs carry identical features and identical labels. One failed run
stands in the archive where a successful run stands in the current cohort. That single substitution
shifts the row order, so the two label arrays disagree at $241$ positions before run identities are
matched. We keep both rows as historical checks and label them here rather than present all six as
one current analysis. On the current caches with corrected task identifiers, the web increment under
grouped folds is $-0.014$ ($[-0.020, -0.009]$), negative as in the archived row, and tau2-bench, with
one run per task, needs no grouped rerun. Figure~\ref{fig:within}(b) reports both policies. Every grouped fold has zero
train-test task overlap, and every fold set partitions the corpus.

Three corpora carry information about leakage, tau-bench, SWE-agent, web, because their runs repeat tasks. Two,
SWE-Gym and tau2-bench, have one run per task, so both policies already hold out whole tasks. The
two algorithms can still choose different folds, and on SWE-Gym they do, but neither corpus tests
sensitivity to repeated-task leakage, so their agreement certifies nothing about it; OpenHands is between, with $26$ of its $574$ tasks holding two runs. SWE-agent is the case
that matters. Its $600$ runs come from $35$ tasks, and grouping removes the lift outright: $+0.085$
becomes $-0.011$, with the interval spanning zero and one positive seed of five. tau-bench keeps a
gain that clears zero at a smaller size. The web corpus is negative under both policies and stays
negative.

The reading is that the SWE-agent lift of \secref{sec:signal} is not established once task identity
is held out, that the tau-bench lift survives at roughly $60\%$ of its run-level size, and that
SWE-Gym's largest-in-the-paper gain is untested by this check rather than confirmed by it. The
intervals here describe split sensitivity on fixed corpora; they do not resample tasks. Resampling
tasks instead ($2000$ draws, pooled out-of-fold predictions, run-level folds) gives
$[-0.005, +0.100]$ on tau-bench, $[+0.084, +0.217]$ on SWE-Gym, $[-0.015, +0.004]$ on OpenHands,
$[-0.028, -0.003]$ on tau2-bench and $[-0.024, +0.005]$ on web; SWE-agent, with $5$ of its $35$ tasks
holding a resolved run, has too few task clusters for that interval.

\subsection{The Ordering Is Coupled to the Baseline}
\label{app:ordering}

A corpus's place in this
order is fixed by its \textsc{flat} baseline, and each gain subtracts that same baseline, so the
two quantities are mathematically coupled. Part of any association between them is therefore
arithmetic. Across the \NObsCorpora{} corpora the association measures Pearson $-0.658$,
Spearman $-0.714$ and Kendall $-0.467$. Four of the 15 corpus pairs put the larger gain on the
higher \textsc{flat} baseline, so the within-corpus ordering falls short of a strict reversal.
A permutation diagnostic that holds each baseline fixed and reassigns the combined scores across
corpora puts the observed association well inside its null, at two-sided $p=0.56$ for Pearson and
$p=0.79$ for Kendall over all $720$ assignments. The ordering is therefore compatible with the
arithmetic alone. That result does not show the dependency effect is absent; it shows this design
and these six points cannot separate the two. The transfer panel of
\appref{app:d} raises the same objection, reaches Kendall $-1.000$ on its logistic arm, and
reports no calibrated test either.

\section{Transfer and the Misread Apparatus}
\label{app:misread}

\subsection{The Full-History Swap, Per Corpus}

\textbf{We can build the degenerate regime and watch the signal collapse.} For each observed-dependency corpus we replace the recovered dependency layer with the full-history layer $\mathcal{A}_0$ on the same steps, compute the \emph{same} size-normalized features on both, and score them under the probe of \secref{sec:signal}. The feature formulas, the flat counts, the labels and the protocol are held fixed. What the replacement varies is the whole construction, so edge endpoints, density and span move together with the attachment grade, and the gap between the columns is the effect of that construction rather than of the grade alone.

In the within-corpus block of Table~\ref{tab:gate} the inferred features fail to improve on flat on
five of \NObsCorpora{} corpora while the observed features lift the three size-weak ones; the lone
exception, SWE-Gym at $0.828$, is run size in disguise (the saturated shape features are nonlinear
functions of run length), and transfer strips the disguise off. In transfer the inferred features
invert where the observed features hold: they follow run size into its two inversions (tau-bench,
SWE-Gym) and add a third, collapsing to $0.287$ on the web corpus where run size alone transfers best
at $0.766$. Counted by held-out corpora below chance, the inferred layer inverts on three of
\NObsCorpora{} where run size inverts on two, transferring worse than the size it encodes.
The chain that reads nothing (\textsc{n0}) clears chance on only \NZeroTrAbove{} of the
\NObsCorpora{} held-out corpora (Table~\ref{tab:n2transfer}). Within corpus the two score alike on SWE-Gym ($0.827$ and $0.828$), and
the reason is structural. Given the four counts, the chain and the full-history layer each add one
direction, $1/n$: every chain feature is $1-1/n$ or $1/n$, and every full-history feature is $1-1/n$
or $(n-1)/2$. The probe therefore sees one model class in both, the counts plus $1/n$.

\subsection{The Control Ladder in Full}
\label{app:ladder}

Figure~\ref{fig:ladderci}, in \secref{subsec:ladder}, gives the estimand
$d = \textsc{aware} - \textsc{n2}$ of Table~\ref{tab:ladder} with its seed-block $95\%$ interval,
its permutation $p$ against the degree-matched null, and the $\textsc{aware} - \textsc{n0}$ column
that H2 was stated on. H1 predicted $d > 0$; the six-corpus mean is
$-0.0003$, and the best corpus, tau-bench, reaches $d = +0.011$ at $p = 0.077$, which becomes
$p = 0.090$ at a raised permutation budget and $0.54$ under Holm across \NObsCorpora{} corpora
(Figure~\ref{fig:ladderci}).
H2 predicted $\textsc{aware} - \textsc{n0} > +0.02$ on a majority; it clears that on two corpora of
\NObsCorpora{} ($+0.048$ and $+0.118$), with a mean of $+0.022$.
A chain that reads nothing beats the recovered layer on three corpora, SWE-Gym ($0.827$ against $0.804$), OpenHands ($0.698$ against $0.695$) and tau2-bench ($0.715$ against $0.706$), while \textsc{aware} wins on the other three.
 The
permutation budget is $25$ draws per corpus, raised to $200$ on tau-bench, the only corpus whose
interval lies above zero. At that budget tau-bench gives $d = +0.010$ with $p = 0.090$, which is
$0.54$ after Holm correction across \NObsCorpora{} corpora. Holm was stated in advance. At $25$
draws the smallest attainable $p$ is $1/26$, above the first Holm threshold of $0.05/6$, so only
the corpus run at $200$ draws could have cleared Holm.

Two figures for tau-bench must not be conflated. The $+0.010$ above is the full-population result.
A separately computed analysis on the $541$ runs carrying both an observed and an inferred grade
gives $d = +0.000$ at $p = 0.423$. A run carries both grades here when it has at least one observed
dependency edge, since the full-history layer over the same steps then adds inferred edges the
recovered layer lacks; the other $119$ runs have none, the degree-matched counterfeit reproduces their
all-zero features, and no inferred edge enters either analysis. These are different populations at different
permutation budgets, and either may be reported with its own label.

\textbf{Counterfeit constructions.} \textsc{aware} is the recovered assignment. \textsc{n1} to \textsc{n3} also hold the edge count fixed, on every one of the
\NLadderRuns{} runs. \textsc{n0} instead attaches each step to its immediate predecessor, reading no
logged information at all, so it builds $n-1$ edges and carries \NZeroEdgeRatio{} times the observed
count. \textsc{n1} draws distinct pairs uniformly. \textsc{n2} preserves each step's number of
earlier sources exactly, its in-degree under \appref{app:formal}'s orientation, while its number of
later consumers varies (one draw in Figure~\ref{fig:twolayer}C). \textsc{n3} preserves that in-degree and fills by recency.

\subsection{Pairwise Versus Pooled Transfer}

\textbf{Pairwise transfer measures a pair.} The naive design trains on one corpus and tests on another, across every ordered pair. Its result is weak: pairwise A-to-B transfer lands at or below chance for most pairs. Two features of the design contribute to that outcome. First, a single corpus is a thin and idiosyncratic training set, so the fitted model carries whatever is particular to that one corpus into the test. Second, a pairwise score mixes the question we care about, whether dependency structure carries across corpora, with one we do not, whether two specific corpora resemble each other. A weak pairwise number is therefore hard to read as evidence about portability in general.

\textbf{Pooled leave-one-corpus-out measures a mixture.} The pooled design trains on five corpora at once and tests on the held-out sixth, rotating the held-out corpus through all \NObsCorpora{}. We use it because the intended training setting holds several corpora at once, and because five pooled corpora make a thicker training set than any one of them. Pooling does not force the model onto signal common to all five. A feature that predicts failure in one training corpus and is uninformative in the other four can keep a positive pooled coefficient, so the other four do not outvote it. Each score evaluates transfer to a corpus held out of training. Five of the six targets keep another corpus from the same broad task family in the training pool, and web is the only held-out family in this protocol. These scores therefore measure corpus-level portability rather than leave-one-family-out transfer across tool use, coding, and web navigation. Which structure produces that portability is left open by these scores.

\subsection{Exploratory Pooled Transfer Panel}
\label{app:d}

\begin{figure*}[t]
\centering
\includegraphics[width=\textwidth]{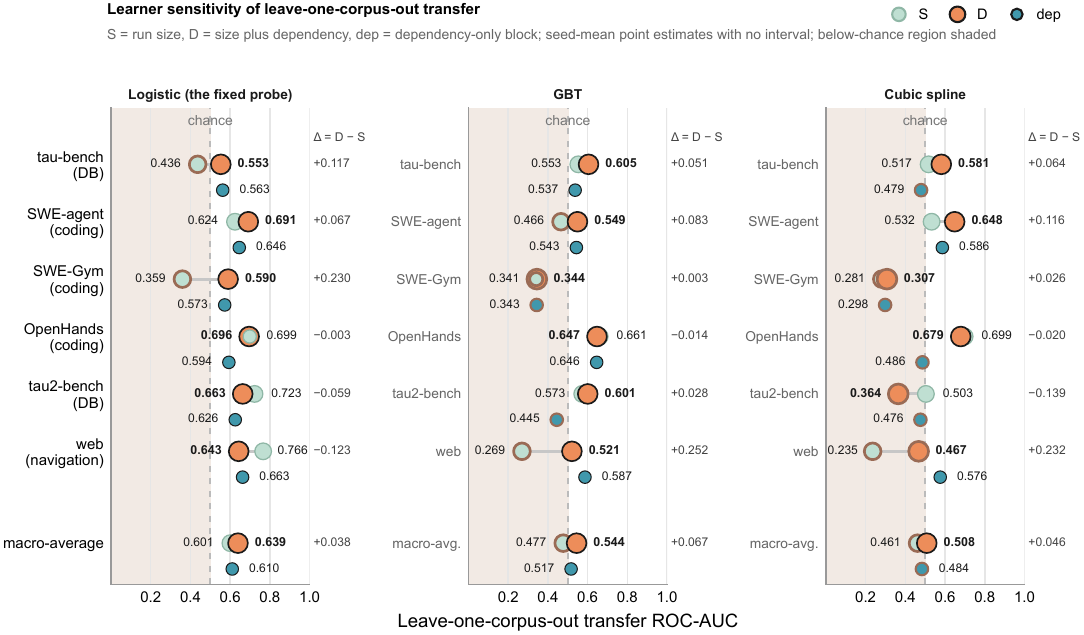}
\Description{Three panels, one per learner, each a horizontal dot plot with one row per corpus and a summary row showing the size-only arm, the size-plus-dependency arm joined to it by a connector, and the dependency-only block, with the region below chance shaded.}
\caption{Leave-one-corpus-out transfer ROC-AUC under three learners, on the shared splits and learner configurations of \appref{app:c1}. $S$ is size alone, $D$ is size plus dependency, and dep is the size-normalized dependency block alone, with no run-size counts, on the same pooled splits. $S$ and $D$ are joined by a gray connector, and the smaller dep marker sits just below its corpus row. The logistic dep values are the \textsc{dep} column of Table~\ref{tab:transferci}. Every marker is a seed-mean point estimate without an uncertainty interval; the shaded region is below chance, and a marker there carries a brown ring. The macro-average row is the unweighted mean over the six targets, computed before rounding, and the $\Delta$ column prints $D-S$ computed before rounding. The per-learner summaries ($\Delta>0$ counts, $S$ and $D$ below-chance counts, ranges and Kendall $\tau$) are in Table~\ref{tab:d-panel}.}
\label{fig:learners}
\end{figure*}

For each of the six target corpora, we train on the pooled runs from the other five and evaluate ROC-AUC on the held-out target, with failure labeled 1. Figure~\ref{fig:learners} and Table~\ref{tab:d-panel} use the same size and dependency blocks, fixed learner configurations, and corrected cache as \appref{app:c1}. All transforms, including spline knots and scaling, are fitted on the training pool. This panel uses the shared web cohort of the primary analysis. Comparisons with earlier results that used another web cohort would combine cohort changes with the dependency producer correction.

\begin{table*}[t]
\centering
\small
\setlength{\tabcolsep}{4pt}
\begin{tabular}{lcccccc}
\toprule
Learner & $\Delta>0$ & AUC${}<0.5$ ($S/D$) & Range $S$ & Range $D$ & Ratio & $\tau$ \\
\midrule
Logistic & 3/6 & 2/0 & $0.407$ & $0.143$ & $2.85$ & $-1.000$ \\
GBT & 5/6 & 3/1 & $0.392$ & $0.303$ & $1.29$ & $-0.600$ \\
Spline & 4/6 & 2/3 & $0.463$ & $0.372$ & $1.25$ & $-0.200$ \\
\bottomrule
\end{tabular}
\caption{Per-learner summaries of the transfer panel in Figure~\ref{fig:learners}. $S$ is size alone, $D$ is size plus dependency, and $\Delta=D-S$ within a target, computed before rounding. Range is maximum minus minimum across these six targets; ratio is range$(S)$/range$(D)$. $\tau$ is Kendall correlation between $S$ and $\Delta$. Every entry is a point summary without an uncertainty interval.}
\label{tab:d-panel}
\end{table*}

The saved computation repeats each fit for seeds 0 to 4. Training and test rows remain fixed, and the seed enters only through the estimator factory. The logistic and spline factories ignore it; this GBT configuration produces identical predictions across seeds on the fixed cache. All 36 cells have exactly zero AUC spread and identical predictions across the five repeats. Their average supplies no variability estimate.

No interval is available in this panel for a paired difference, either arm's macro-average, the macro-average difference, a minimum, the difference of minima, a range ratio, or Kendall correlation. The separate gating analysis computes marginal held-out-run bootstrap intervals for individual logistic arms on the same target runs. Subtracting the saved marginal interval endpoints cannot construct a paired interval. The current summaries also do not quantify uncertainty from training data or variation across a population of corpora.

Logistic regression has three positive and three negative target differences, with a macro-average difference of $+0.038$. Its dependency-arm AUCs are all above $0.5$ as point estimates. The observed ranges shrink from $0.407$ to $0.143$, a ratio of $2.85$. Ordering targets by the logistic size AUC places their differences in reverse order, with all 15 target pairs concordant with decreasing $\Delta$ and $\tau=-1.000$. This ordering, range comparison, and earlier emphasis on minima followed inspection of the six targets. Because $\Delta$ subtracts $S$, the plotted quantities are mathematically coupled; no calibrated test of that relationship is reported. Neither the ordering nor the smaller range identifies a mechanism or predicts performance on an unseen corpus.

We omit the difference between separately minimized arms. For logistic, the size minimum occurs on SWE-Gym and the augmented minimum on tau-bench, so their difference describes no target's paired change.

Learner sensitivity changes both the signs and the ordering. On web, the difference is $-0.123$ for logistic and $+0.252$ for GBT. On tau2-bench, it is $+0.028$ for GBT and $-0.139$ for spline. The number of dependency-arm point AUCs below $0.5$ is zero for logistic, one for GBT, and three for spline.

\textbf{The dependency-only block under the same three learners.} The dep markers of
Figure~\ref{fig:learners} refit the size-normalized dependency block alone, with no run-size
counts, on the same pooled splits and with the same learner configurations. Under logistic
regression every target sits above chance, which is the transfer sentence of the abstract. Under
GBT two targets fall below chance and under the cubic spline four do, against one and three for
size plus dependency in the same figure. The dependency-only arm is therefore the more
learner-sensitive of the two. Every marker in Figure~\ref{fig:learners} is a seed-mean point
estimate with no uncertainty interval.

\begin{table}[t]
\centering\footnotesize
\setlength{\tabcolsep}{3pt}
\caption{Task-level 95\% intervals for the transfer AUCs of Table~\ref{tab:transferci}, from $2000$
paired draws of held-out tasks that condition on the fitted models. SWE-agent has too few task
clusters for this interval. On the same draws the paired differences from \textsc{flat} have
intervals $[+0.025, +0.219]$ (\textsc{dep}) and $[+0.006, +0.223]$ (\textsc{f{+}d}) on tau-bench,
$[+0.150, +0.280]$ and $[+0.172, +0.292]$ on SWE-Gym, and $[-0.145, -0.063]$ and $[-0.007, +0.001]$ on
OpenHands; on tau2-bench and web both lie below zero.}
\label{tab:transfertask}
\begin{tabular}{@{}lccc@{}}
\toprule
Held-out & \textsc{flat} & \textsc{dep} & \textsc{f{+}d} \\
\midrule
tau-bench  & [0.382, 0.495] & [0.496, 0.626] & [0.483, 0.620] \\
SWE-agent  & --- & --- & --- \\
SWE-Gym    & [0.301, 0.417] & [0.513, 0.631] & [0.530, 0.647] \\
OpenHands  & [0.659, 0.740] & [0.547, 0.639] & [0.655, 0.738] \\
tau2-bench & [0.661, 0.779] & [0.560, 0.694] & [0.601, 0.726] \\
web        & [0.717, 0.811] & [0.607, 0.716] & [0.585, 0.697] \\
\bottomrule
\end{tabular}
\end{table}

\subsection{Transfer Uncertainty}
\label{app:transferci}

Table~\ref{tab:transferci} attaches a confidence interval to every leave-one-corpus-out transfer
AUC. The pooled fit is deterministic, so the uncertainty that matters is sampling of the held-out
corpus: we resample the held-out runs with replacement ($2000$ draws) for the 95\% interval. We
report the intervals and do not attach a significance test to them, because six held-out corpora
screened one at a time is a description of where the signal holds rather than a hypothesis under
test. This table and Table~\ref{tab:gate} are cut from a single run of the gating experiment, so
every point AUC the two share is the same number. Every \textsc{dep} and \textsc{f{+}d} transfer
AUC sits above $0.5$ with its whole interval above the chance line; the two inverted \textsc{flat}
entries (tau-bench, SWE-Gym) sit below it. Resampling held-out tasks rather than runs widens the
intervals on tau-bench and web, where tasks repeat up to four times (Table~\ref{tab:transfertask}):
both tau-bench intervals then include $0.5$, and SWE-agent, with $5$ of its $35$ tasks holding a
resolved run, has too few task clusters for a task-level interval. The task intervals use $2000$
paired draws of held-out tasks and condition on the fitted models.

\begin{table}[t]
\centering\small
\setlength{\tabcolsep}{4pt}
\caption{Leave-one-corpus-out transfer AUC with a held-out bootstrap 95\% CI, for run size
(\textsc{flat}), the size-normalized dependency layer (\textsc{dep}), and flat plus that layer
(\textsc{f{+}d}). \textit{Italic} marks a below-chance (inverted) AUC. No significance test is
attached; see \appref{app:transferci}.}
\label{tab:transferci}
\begin{tabular}{@{}llcc@{}}
\toprule
Held-out & Features & AUC & 95\% CI \\
\midrule
tau-bench  & \textsc{flat}  & \textit{0.436} & [0.392, 0.481] \\
           & \textsc{dep}   & 0.563 & [0.520, 0.607] \\
           & \textsc{f{+}d} & 0.553 & [0.509, 0.596] \\
\midrule
SWE-agent  & \textsc{flat}  & 0.624 & [0.559, 0.690] \\
           & \textsc{dep}   & 0.646 & [0.574, 0.716] \\
           & \textsc{f{+}d} & 0.691 & [0.624, 0.757] \\
\midrule
SWE-Gym    & \textsc{flat}  & \textit{0.359} & [0.297, 0.420] \\
           & \textsc{dep}   & 0.573 & [0.512, 0.633] \\
           & \textsc{f{+}d} & 0.590 & [0.532, 0.648] \\
\midrule
OpenHands  & \textsc{flat}  & 0.699 & [0.655, 0.740] \\
           & \textsc{dep}   & 0.594 & [0.548, 0.637] \\
           & \textsc{f{+}d} & 0.696 & [0.652, 0.738] \\
\midrule
tau2-bench & \textsc{flat}  & 0.723 & [0.662, 0.781] \\
           & \textsc{dep}   & 0.626 & [0.561, 0.689] \\
           & \textsc{f{+}d} & 0.663 & [0.602, 0.725] \\
\midrule
web        & \textsc{flat}  & 0.766 & [0.728, 0.806] \\
           & \textsc{dep}   & 0.663 & [0.619, 0.708] \\
           & \textsc{f{+}d} & 0.643 & [0.598, 0.688] \\
\bottomrule
\end{tabular}
\end{table}

\subsection{The Control Ladder, Scored in Transfer}
\label{app:n2transfer}

\secref{subsec:ladder} scores the attachment ladder within corpus. Table~\ref{tab:n2transfer} scores
the same four rules in the pooled leave-one-corpus-out design of \appref{app:transferci}. The feature
definitions and the attachment rules are unchanged, and the evaluation is what differs. Each null
block replaces the dependency columns on the
training corpora and on the held-out corpus alike, since the arm is the whole pipeline rather than a
perturbation of the test set. \textsc{n1} and \textsc{n2} draw \NTwoTrDraws{} times here and we
report the draw mean; \textsc{n0} and
\textsc{n3} are deterministic and drawn once. Two details keep the two panels from being read as
paired. This run shares one generator across corpora for each rule and draw, while the within-corpus
ladder resets inside each corpus, so the later corpora see different draws. That ladder also used
$25$ draws for its six-corpus mean rather than \NTwoTrDraws{}. Rerunning this panel under the
ladder's reset convention moves the six-corpus mean from \NTwoTrMeanGap{} to \NTwoTrResetGap{} and
changes no sign, no above-chance count and not the Holm survivor. The contrast below is therefore
between two evaluation settings rather than an estimate of what switching the convention does. Two gates ran before any null was scored. The
dependency columns recomputed from the ladder's own code are bit-identical to this panel's cached
features on every corpus, and all twelve saved \textsc{flat} and \textsc{flat+dep} transfer cells
reproduce exactly, including each of the sixty per-seed values. The protocol behind this panel was
registered after the \textsc{aware} transfer values existed and before either gate or any null arm
ran. Its primary rule counts the held-out corpora on which an arm scores above chance (the last row of
Table~\ref{tab:n2transfer}). Outcomes written in advance treat \textsc{aware}'s finding, above
chance on every held-out corpus, as surviving the control if \textsc{n2} falls below chance on some
targets or has materially lower AUC. They call for a narrower claim if \textsc{n2} also clears
chance on every target at similar AUC, and report a partial pattern as mixed. No materiality
threshold was set. The run's report assigns the observed pattern, \textsc{n2} below chance on
SWE-Gym alone, to the survival outcome, so that verdict rests on a single corpus.

\begin{table*}[t]
\centering\small
\setlength{\tabcolsep}{5pt}
\begin{tabular}{lrrrrrrcrr}
\toprule
Held-out & \textsc{flat} & \textsc{aware} & \textsc{n0} & \textsc{n1} & \textsc{n2} & \textsc{n3}
  & \textsc{n2} draw range & $\textsc{aware}-\textsc{n2}$ & $p$ \\
\midrule
tau-bench  & \textit{0.436} & 0.553 & \textit{0.443} & 0.545 & 0.536 & \textit{0.496} & $0.518$--$0.564$ & $+0.017$ & $0.020$ \\
SWE-agent  & 0.624 & 0.691 & 0.605 & 0.690 & 0.689 & 0.691 & $0.677$--$0.704$ & $+0.002$ & $0.348$ \\
SWE-Gym    & \textit{0.359} & 0.590 & \textit{0.342} & \textit{0.416} & \textit{0.481} & 0.531 & $0.431$--$0.529$ & $+0.109$ & $0.005$ \\
OpenHands  & 0.699 & 0.696 & 0.699 & 0.698 & 0.697 & 0.697 & $0.695$--$0.698$ & $-0.002$ & $0.995$ \\
tau2-bench & 0.723 & 0.663 & 0.723 & 0.667 & 0.668 & 0.670 & $0.640$--$0.685$ & $-0.005$ & $0.791$ \\
web        & 0.766 & 0.643 & \textit{0.287} & 0.534 & 0.594 & 0.645 & $0.538$--$0.656$ & $+0.049$ & $0.020$ \\
\midrule
Above chance & 4 & 6 & 3 & 5 & 5 & 5 & & mean $+0.028$ & \\
\bottomrule
\end{tabular}
\caption{\textbf{The attachment ladder in transfer.} Pooled leave-one-corpus-out AUC under each
attachment rule, with the same fixed logistic probe as Table~\ref{tab:transferci}. The
\textsc{flat} and \textsc{aware} columns are that table's \textsc{flat} and \textsc{f{+}d} entries.
Italic marks an entry below $0.5$. $p$ is the fraction of \NTwoTrDraws{} degree-matched draws
reaching or exceeding \textsc{aware} on that corpus, by the ladder's $(1+\text{ge})/(1+B)$ convention;
it is one-sided, so a value near $1$ means the counterfeit scored higher. Under Holm across
\NObsCorpora{} corpora only SWE-Gym clears $0.05$. The five seed calls return identical AUCs, so the
seed-block half-width is zero on every cell. The \textsc{n2} draw range is the minimum and maximum
over the \NTwoTrDraws{} draws; it describes attachment randomization on fixed runs. This panel
carries no sampling interval for the individual arms, and the bootstrap intervals of
Table~\ref{tab:transferci} cannot be borrowed for a paired contrast.
\appref{app:n2transfer} gives a paired interval for the $\textsc{aware}-\textsc{n2}$ column.}
\label{tab:n2transfer}
\end{table*}

Three things separate this from the within-corpus ladder. Its estimand changes sign and size, from
$-0.0003$ within corpus to \NTwoTrMeanGap{} here. The control gives nominal separations on
\NTwoTrSeparates{} corpora here and on none within corpus, and only SWE-Gym clears Holm in the $p$
column. No draw reached \textsc{aware} on SWE-Gym ($p = \NTwoTrSweGymP{}$), and some draws reached it on tau-bench and web ($p = \NTwoTrTauWebP{}$). And \textsc{n2} loses the all-corpora-above-chance pattern that \textsc{aware} keeps,
failing on SWE-Gym at $0.481$. Against that, the counterfeit scores higher on \NTwoTrInverts{}
corpora, and every arm that keeps each step's number of earlier sources reaches \NTwoTrAbove{} of
\NObsCorpora{} on its own. A universal advantage for the recovered
attachment is therefore false here. In the attachment-randomization screen, the pipeline transfers
better with the recovered endpoints on SWE-Gym after Holm correction and nominally on tau-bench and
web. The paired bootstrap was registered after every entry of Table~\ref{tab:n2transfer} existed,
its $p$ column and Holm survivor included. Its protocol quotes some of those values and names
SWE-Gym as the corpus the attachment screen selected. The resampling unit, replicate count, Holm
family and outcome branches, including a branch that would have retracted the SWE-Gym result, were fixed
before any bootstrap quantity was computed. Resampling the held-out corpus $2000$ times,
clustered by task, gives paired 95\% intervals of $\NTwoCiSweGym$ on SWE-Gym and $\NTwoCiMean$
on the six-corpus mean. One resample is shared by \textsc{aware} and all \NTwoTrDraws{}
\textsc{n2} blocks, which is what makes the interval paired. These intervals resample held-out tasks
while conditioning on the fitted models, the five training corpora and the 200 realized
\textsc{n2} draws; they quantify evaluation-unit sampling within these six corpora.
Table~\ref{tab:pairedci} carries all \NObsCorpora{} of them with the run-level panel beside each.
The same three corpora that separate nominally in the attachment screen
clear Holm across the six paired task-bootstrap tests.
The printed 95\% intervals are pointwise.
The paired tests concern held-out task sampling, whereas the $p$ column
concerns attachment randomization, so their Holm outcomes are not interchangeable. Resampling runs instead, as
\appref{app:transferci} does, changes no decision.

\begin{table*}[t]
\centering\small
\setlength{\tabcolsep}{6pt}
\begin{tabular}{lrccrl}
\toprule
Held-out & $\textsc{aware}-\textsc{n2}$ & Task $95\%$ CI & Run $95\%$ CI & $p$ (task) & Holm (task) \\
\midrule
tau-bench  & $+0.017$ & $[+0.005, +0.028]$ & $[+0.007, +0.027]$ & $0.006$ & survives \\
SWE-agent  & $+0.002$ & $[-0.012, +0.015]$ & $[-0.011, +0.016]$ & $0.730$ & no \\
SWE-Gym    & $+0.109$ & $[+0.086, +0.133]$ & $[+0.085, +0.134]$ & $0.001$ & survives \\
OpenHands  & $-0.002$ & $[-0.005, +0.002]$ & $[-0.005, +0.002]$ & $0.337$ & no \\
tau2-bench & $-0.005$ & $[-0.019, +0.010]$ & $[-0.018, +0.010]$ & $0.510$ & no \\
web        & $+0.049$ & $[+0.019, +0.080]$ & $[+0.024, +0.073]$ & $0.002$ & survives \\
\midrule
Six-corpus mean & $+0.028$ & $[+0.021, +0.036]$ & $[+0.022, +0.035]$ & $0.001$ & not corrected \\
\bottomrule
\end{tabular}
\caption{\textbf{The paired held-out interval on the $\textsc{aware}-\textsc{n2}$ column of
Table~\ref{tab:n2transfer}.} Percentile $95\%$ intervals from $2000$ bootstrap replicates of the
held-out corpus. One resample is shared by \textsc{aware} and all \NTwoTrDraws{} \textsc{n2} draws,
which is what pairs the two arms. The task panel is primary and clusters by task; the run panel
resamples runs, matching Table~\ref{tab:transferci}. $p$ is the two-sided bootstrap tail of the task
panel under the plus-one convention. Holm runs at family-wise $0.05$ across these \NObsCorpora{}
per-corpus tests, and the run panel returns the same \NTwoTrSeparates{} survivors. That family
covers sampling of evaluation units, with the fitted models and the \NTwoTrDraws{} realized draws
held fixed. It is therefore separate from the attachment-randomization screen of
Table~\ref{tab:n2transfer}, whose $p$ column leaves SWE-Gym alone. The six-corpus mean is one
prespecified summary and takes no correction. SWE-Gym and tau2-bench hold one run per task, so the
two units coincide there and the small endpoint differences come from the two independent random
streams. Six replicates on SWE-agent drew a single label class and were recorded as dropped rather
than redrawn. The joint mean keeps the $1{,}994$ replicates in which all \NObsCorpora{} corpora
carry both classes.}
\label{tab:pairedci}
\end{table*}

\subsection{Withholding a Whole Task Family}
\label{app:familytransfer}

\appref{app:transferci} and \appref{app:n2transfer} hold out one corpus at a time, and every target
except web keeps a corpus from its own broad task family in the training pool. This subsection
withholds the family. The \FamFamilies{} families partition the \NObsCorpora{} corpora, frozen before any
quantity of this panel existed: coding (SWE-agent, SWE-Gym, OpenHands), database and tool interaction
(tau-bench, tau2-bench), and web (AgentRewardBench). For a target corpus the training pool is every
corpus whose family differs from its own. The three coding targets therefore train on tau-bench,
tau2-bench and web ($1{,}538$ runs); the two database and tool targets train on SWE-agent, SWE-Gym,
OpenHands and web ($2{,}176$ runs); web trains on the other \FamNewTargets{} ($2{,}514$ runs).
Everything else is the pipeline of \appref{app:n2transfer}: the same caches, the same fixed logistic
probe, the same \NTwoTrDraws{} realized degree-matched draws, and the same $2{,}000$ paired
replicates with one index vector per replicate shared by \textsc{aware} and all \NTwoTrDraws{}
\textsc{n2} blocks.

\textbf{Web is reused evidence, and carries $\dagger$ wherever it appears.} Its family holds one
corpus, so withholding that family removes nothing beyond the target itself, and web's pool is its
leave-one-corpus-out pool. Every web entry in Tables~\ref{tab:famtransfer} and~\ref{tab:famboot} is
the published leave-one-corpus-out value reprinted for context, and a gate required that
reproduction and measured a maximum absolute difference of $0.0$ over every web quantity. Web is
therefore in no Holm family, no survivor count and no summary, and it replicates nothing here.

\begin{table*}[t]
\centering\footnotesize
\setlength{\tabcolsep}{4pt}
\begin{tabular}{lrrrrcrrrl}
\toprule
 & & \multicolumn{5}{c}{Family-withheld transfer AUC}
  & \multicolumn{3}{c}{Attachment screen (family S, $m=5$)} \\
\cmidrule(lr){3-7}\cmidrule(lr){8-10}
Held-out & Runs & \textsc{flat} & \textsc{aware} & \textsc{n2} mean & \textsc{n2} range
  & $\textsc{aware}-\textsc{n2}$ & $\geq$ \textsc{aware} & $p$ & Holm \\
\midrule
\multicolumn{10}{@{}l}{\textit{coding} withheld; pool tau-bench, tau2-bench, web ($1{,}538$ runs)} \\
SWE-agent$^{\ddagger}$ & 600 & 0.624 & 0.633 & 0.636 & $0.629$--$0.644$ & $-0.003$ & 162/200 & 0.811 & 1.000, no \\
SWE-Gym    & 376 & \textit{0.339} & \textit{0.343} & \textit{0.379} & $0.343$--$0.423$ & $-0.036$ & 199/200 & 0.995 & 1.000, no \\
OpenHands  & 600 & 0.699 & 0.698 & 0.699 & $0.697$--$0.700$ & $-0.001$ & 166/200 & 0.831 & 1.000, no \\
\midrule
\multicolumn{10}{@{}l}{\textit{database and tool} withheld; pool SWE-agent, SWE-Gym, OpenHands, web ($2{,}176$ runs)} \\
tau-bench  & 660 & \textit{0.423} & 0.518 & 0.516 & $0.503$--$0.534$ & $+0.002$ & 66/200 & 0.333 & 1.000, no \\
tau2-bench & 278 & \textit{0.291} & 0.548 & 0.577 & $0.542$--$0.601$ & $-0.029$ & 198/200 & 0.990 & 1.000, no \\
\midrule
\multicolumn{10}{@{}l}{\textit{web} withheld; pool the other \FamNewTargets{} corpora ($2{,}514$ runs), unchanged from leave-one-corpus-out} \\
web$^{\dagger}$ & 600 & 0.766 & 0.643 & 0.594 & $0.538$--$0.656$ & $+0.049$ & 3/200 & 0.020 & excluded \\
\bottomrule
\end{tabular}
\caption{\textbf{The family-withheld transfer panel and its attachment screen.} Pooled AUC under a
training pool holding no corpus from the target's own frozen family, with the features, probe and
\NTwoTrDraws{} realized draws of Table~\ref{tab:n2transfer}. Italic marks an entry below $0.5$.
\textsc{n2} range is the minimum and maximum over the \NTwoTrDraws{} draws, and $\geq$ \textsc{aware}
counts the draws reaching or exceeding it. The screen's $p$ is that count under the
$(1+\text{ge})/(1+B)$ convention; it is one-sided, so a value near $1$ means the counterfeit scored
higher. Holm runs at family-wise $0.05$ over the attachment screen's own family S, whose
\FamNewTargets{} members are the targets whose pool changed, and it leaves no survivor; no new target
reaches $0.05$ before correction either. $\dagger$ reused evidence: web's pool is unchanged from the
leave-one-corpus-out panel, so this row reprints published values and is not an independent
replication. $\ddagger$ SWE-agent's task-unit result was flagged uninterpretable at registration; it
stays in this screen, whose unit is the attachment draw rather than the task cluster.}
\label{tab:famtransfer}
\end{table*}

\begin{table*}[t]
\centering\footnotesize
\setlength{\tabcolsep}{3pt}
\begin{tabular}{lrcrlcrl}
\toprule
 & & \multicolumn{3}{c}{Task unit, primary (family B-task, $m=4$)}
  & \multicolumn{3}{c}{Run unit, secondary (family B-run, $m=5$)} \\
\cmidrule(lr){3-5}\cmidrule(lr){6-8}
Held-out & $\textsc{aware}-\textsc{n2}$ & $95\%$ CI & $p$ & Holm & $95\%$ CI & $p$ & Holm \\
\midrule
tau-bench  & $+0.002$ & $[-0.030, +0.035]$ & 0.917 & 1.000, no & $[-0.023, +0.026]$ & 0.909 & 1.000, no \\
SWE-agent$^{\ddagger}$ & $-0.003$ & $[-0.011, +0.004]$ & 0.521 & excluded & $[-0.006, +0.001]$ & 0.164 & 0.492, no \\
SWE-Gym    & $-0.036$ & $[-0.048, -0.025]$ & 0.001 & 0.004 survives & $[-0.048, -0.025]$ & 0.001 & 0.005 survives \\
OpenHands  & $-0.001$ & $[-0.003, +0.001]$ & 0.556 & 1.000, no & $[-0.003, +0.001]$ & 0.558 & 1.000, no \\
tau2-bench & $-0.029$ & $[-0.056, -0.001]$ & 0.045 & 0.135, no & $[-0.057, -0.003]$ & 0.029 & 0.116, no \\
web$^{\dagger}$ & $+0.049$ & $[+0.019, +0.080]$ & 0.002 & excluded & $[+0.024, +0.073]$ & 0.001 & excluded \\
\midrule
S1 mean & $\FamSOne$ & $\FamSOneCI$ & 0.005 & not corrected & & & \\
S2 mean & $\FamSTwo$ & & & & $\FamSTwoCI$ & 0.001 & not corrected \\
\bottomrule
\end{tabular}
\caption{\textbf{The paired held-out bootstrap on the family-withheld panel.} Percentile $95\%$
intervals from $2{,}000$ resamples of the held-out corpus, with one index vector per replicate shared
by \textsc{aware} and all \NTwoTrDraws{} \textsc{n2} draws, which is what pairs the arms. $p$ is the
two-sided bootstrap tail under the plus-one convention. The two units carry separate Holm families,
B-task over four targets and B-run over \FamNewTargets{}, and each leaves SWE-Gym alone and in the
negative direction. Both families cover sampling of the evaluation units with the fits and the
\NTwoTrDraws{} realized draws held fixed, so their outcomes are not interchangeable with the
attachment screen's family S in Table~\ref{tab:famtransfer}. $\dagger$ reused evidence, excluded from
every family and summary. $\ddagger$ SWE-agent's task-unit row was flagged uninterpretable at
registration, because $5$ of its $35$ task clusters carry the minority class against a preregistered
floor of $10$; the row is printed and is not evidence, and it is why B-task holds four members while
B-run holds \FamNewTargets{}. Six of SWE-agent's $2{,}000$ task replicates drew a single label class
and were recorded as dropped; every other cell retains all $2{,}000$. S1 and S2 are the two summaries
declared before the run, both web-excluded, both secondary and uncorrected: S1 averages
$\textsc{aware}-\textsc{n2}$ over the four members of B-task at the task unit and S2 over the
\FamNewTargets{} members of B-run at the run unit, each from a joint resample.}
\label{tab:famboot}
\end{table*}

\textbf{The two tests keep separate families, as they do on the leave-one-corpus-out panel.} The
attachment screen varies which earlier step each dependency edge attaches to and conditions on the
realized evaluation sample; the paired bootstrap resamples the evaluation units and conditions on the
fits and the \NTwoTrDraws{} realized draws. Here the screen's family S holds \FamNewTargets{} members
and leaves no survivor, while the bootstrap's families hold four members at the task unit and
\FamNewTargets{} at the run unit and leave SWE-Gym alone in each, in the negative direction. The
three sets of Holm outcomes answer different questions and are never merged or reported under one
word.

\textbf{Four of the \FamNewTargets{} new targets carry a negative contrast}, and both prespecified
equal-corpus summaries are negative with intervals clear of zero (\FamSOne{}, $\FamSOneCI$ at the
task unit over four corpora; \FamSTwo{}, $\FamSTwoCI$ at the run unit over \FamNewTargets{}). SWE-Gym
carried the leave-one-corpus-out finding and loses the most. A gate reproduced its leave-one-corpus-out
\textsc{aware} AUC at \SweGymTrBoth{} on these same evaluation rows, and the family-withheld pool puts
it at \FamAwareSweGym{}, a move of $\FamSweGymDrop{}$ that lands below chance. \textsc{flat} falls on
the two database and tool targets as well, which says the shrunken mixture carries less of
everything; it is a diagnostic and carries no test here.

\textbf{The protocol was written before any result existed, and a second implementation re-derived
the panel.} The registration fixed the families, the six splits, the two tests, the Holm memberships,
the primary unit, the two summaries and the structural floors while no AUC, draw, replicate or
interval of this panel had been computed, and it recorded that boundary. The re-derivation reads the
same pinned caches under its own code: it matched all $11$ input hashes, regenerated the
\NTwoTrDraws{} degree-matched blocks and reproduced the leave-one-corpus-out cells exactly, and
agreed on every survivor list. Over \FamVerifyN{} compared quantities its maximum absolute
discrepancy is $\FamVerifyMax$, which is floating-point noise.

\textbf{What \FamFamilies{} families cannot settle.} One family holds a single corpus and another
holds two sibling releases of one benchmark, so this panel bounds how far the leave-one-corpus-out
reading travels without settling how the representation behaves across task domains in general, in
either direction. Family withholding also shrinks the training pool and changes its relatedness to
the target at the same time: the coding targets lose $976$ to $1{,}200$ runs and the database and
tool targets lose $278$ and $660$, and no control here separates size from relatedness. A null is
consistent with the representation failing to travel across families, with a three-corpus pool being
too small to fit the probe, or with both. The leave-one-corpus-out result of \appref{app:n2transfer}
was measured under a different training pool and is not re-measured here.

\subsection{Four Ways a Generic Representation Misreads}
\label{app:genericmisread}

The degenerate regime turns into four concrete difficulties when a permutation-invariant graph model meets the two-layer graph. Each is tied to a measurement already in hand.

\begin{enumerate}
\item \textbf{Size- and permutation-blind pooling keys on run size.} A standard sum or mean readout over the degenerate layer recovers a function of $n$, because at step $i$ the full-history degrees are $i-1$ toward earlier steps and $n-i$ toward later ones, and both are fixed by $n$. Run size is the signal \secref{sec:unification} shows is not corpus-invariant: pooled flat features fall below chance on two held-out corpora, as low as $0.36$ on SWE-Gym. A pooled readout therefore locks onto a feature that can invert across corpora.
\item \textbf{No channel for the source label.} The graph carries observed execution edges, declared dependency edges, and inferred dependency edges, but a generic message-passing layer sees one adjacency and cannot condition on the attachment grade $\sigma$. It cannot down-weight an inferred edge relative to a declared one, so the size-collinear inferred layer ($\lvert r \rvert$ up to \MaxInfDepSizeCorr{} when the evaluation corpora are given full-history edges, and $1.00$ on the Who\&When runs of Figure~\ref{fig:degenerate}) is propagated as if it were observed signal. Generic message passing has no inductive bias for this, because the bias that would help, treating declared and inferred adjacency as different relations, is not one it carries by default.
\item \textbf{Transfer performance requires separate evaluation.} In the regenerated five-seed GIN results, pooled transfer AUC is lower than the lean graph features on every held-out corpus (Figure~\ref{fig:gnn}). These model comparisons also change architecture and representation, so they do not isolate capacity as the cause of poorer transfer.
\item \textbf{A joint embedding states the marginal lift only under a nested ablation, and the lift it then reports varies by corpus and model.} A single embedding of the whole graph returns one score. The lift therefore has to be recovered by fitting the same architecture twice on identical splits, once on the execution layer alone and once with the dependency layer added. We ran that nested ablation, so this is a measurement rather than an impossibility. All twelve cells come from one sweep: one device, five seeds, the shared cohort. On tau-bench the relation-aware R-GCN gains $+0.031 \pm 0.010$ from the real dependency layer against $-0.015 \pm 0.019$ from a count-matched rewiring of it, a paired gap of $+0.047$. Over the \NObsCorpora{} corpora the mean lift is $+0.015$ for R-GCN and $-0.008$ for the edge-type-blind GIN, and neither pattern is uniform: the GIN gains $+0.018 \pm 0.003$ on the web corpus and loses $-0.067 \pm 0.009$ on OpenHands, while R-GCN loses on tau2-bench. Against the rewiring control the result is cleaner, the real layer beating it in eleven of twelve cells. Edge-type blindness therefore does not by itself stop a model from responding to dependency endpoints. What a monolithic fit forfeits is reading the decomposition off one score, not the estimand.
\end{enumerate}

The first three are readings of one blind spot: the attachment grade is invisible to a model built for graphs whose edges all mean the same thing. The fourth is a constraint on how a joint model has to be fitted if it is to price the layer at all.

\subsection{Graph Networks, Per Corpus}
\label{app:gin}
\label{subsec:gin}

\begin{figure*}[t]
\centering
\includegraphics[width=\textwidth]{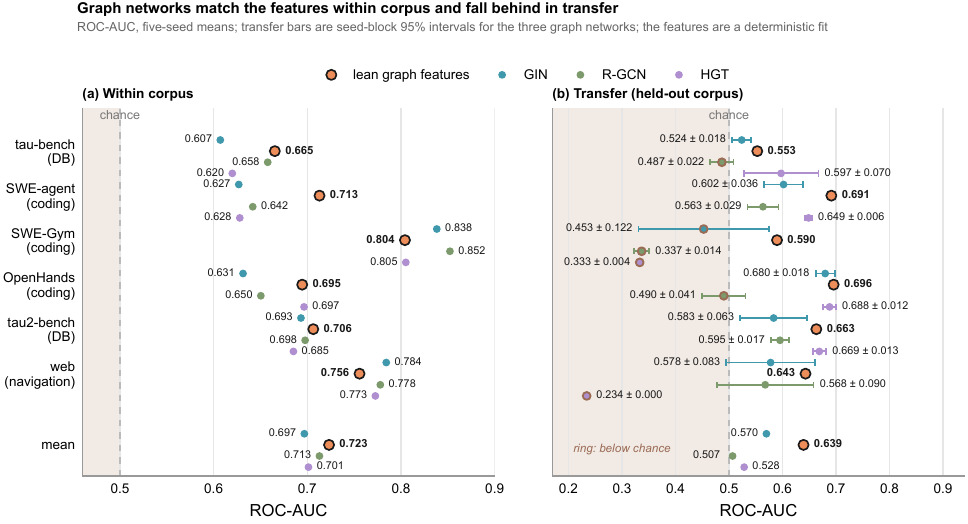}
\Description{Two panels of horizontal dot plots, one row per corpus plus a summary row, comparing the lean graph features with three graph networks, within corpus on the left and in leave-one-corpus-out transfer on the right, where the graph networks carry seed intervals and the region below chance is shaded.}
\caption{The lean graph features against three off-the-shelf graph networks on the full two-layer graph: the edge-type-blind GIN~\cite{xu2019gin}, and the relation-aware R-GCN~\cite{schlichtkrull2018rgcn} and HGT~\cite{hu2020hgt}, which read the edge type. Scores are within corpus and in pooled leave-one-corpus-out transfer (ROC-AUC, label $1$ a failed run, chance $0.5$, mean over five seeds). Within corpus the graph networks are competitive with the features (mean $0.70$ to $0.71$ against $0.72$); in transfer they diverge. The features hold a $0.639$ mean and a $0.553$ worst case, while every graph network is lower on the mean and the two relation-aware models drop below chance on held-out corpora (R-GCN to $0.337$, HGT to $0.234$), and the GIN dips below chance once, on SWE-Gym. At the untuned settings of \appref{app:gin}, reading the edge type buys within-corpus fit without improving transfer to held-out corpora. All eight series come from saved artifacts: the features from the gating run, the graph networks from a single five-seed sweep of all 36 cells. Transfer bars are five-seed 95\% intervals (seed-block $t$, $\mathrm{df}=4$) on the pooled leave-one-corpus-out transfer AUC of the three graph networks. The off-the-shelf models score lower and swing across seeds. The GIN's transfer moves by up to $\pm0.12$, and R-GCN and HGT each cross below chance on held-out corpora. These intervals vary the training seed on a fixed target. The lean graph feature intervals of Table~\ref{tab:transferci} resample held-out runs for a deterministic fit. The two widths therefore answer different questions, and comparing them does not establish a stability advantage.}
\label{fig:gnn}
\end{figure*}

We make the cost concrete with three off-the-shelf networks under the same five-seed, five-fold protocol: the edge-type-blind GIN~\cite{xu2019gin}, and two relation-aware models that read the edge type, R-GCN~\cite{schlichtkrull2018rgcn} and HGT~\cite{hu2020hgt}. The lean graph features are the \textsc{flat+dep} set of \secref{sec:unification} under the logistic probe: the four counts and the size-normalized dependency block. Within corpus all three match the lean graph features (mean AUC $0.70$ to $0.71$ against $0.72$). In pooled transfer the ranking inverts (Figure~\ref{fig:gnn}). The lean graph features hold a $0.639$ mean and never fall below $0.553$. The GIN trails on all six corpora, and the two relation-aware models drop below chance on held-out corpora. HGT is the one partial exception, clearing the features on two corpora while losing the mean by more than a tenth. All three are also unstable, the GIN swinging up to $\pm0.12$ across seeds (Figure~\ref{fig:gnn}(b)).

\textbf{Settings.} Each network has two message-passing layers of width $64$ (HGT with four heads)
over $8$-dimensional node features: a node-type indicator, a read, write or neither access tag, and
the step's normalized position. Edges carry no features. The graphs hold no resource nodes, so the
populated relations are emit, handoff and dependency edges; GIN merges them into one edge set, and
R-GCN and HGT keep them typed. The graph vector concatenates the \texttt{global\_mean\_pool} and
\texttt{global\_max\_pool} outputs of PyTorch Geometric and feeds a two-layer head with dropout $0.3$.
Training runs $40$ epochs of Adam (learning rate $10^{-3}$, weight decay $10^{-4}$, batch $32$)
without hyperparameter tuning or early stopping, weighting the failure class by $(1-p)/p$ for training failure rate $p$.
Seeds $0$ to $4$ set both the initial weights and the batch order. Within corpus each seed runs five
stratified folds; in transfer each seed trains once on the pooled other five corpora. HGT's five
transfer AUCs on the web corpus span $0.2339$ to $0.2347$, so its seed interval has half-width
$0.0004$ and prints as $\pm 0.000$ at three decimals.

\textbf{A message-passing network is fooled the same way.} The edge-type-blind GIN, rerun on the inferred graphs over five seeds, falls below chance on the web corpus ($0.244 \pm 0.007$) and on SWE-Gym ($0.351 \pm 0.095$), fooled by the saturated layer as the linear probe is. Only the web corpus separates the two constructions: there the observed graph stays above chance at $0.567 \pm 0.146$, while on SWE-Gym the observed graph is below chance too, at $0.469 \pm 0.120$. These observed-graph scores come from the rerun paired with the
inferred graphs, a separate run from the sweep in Figure~\ref{fig:gnn}, so they differ slightly
from that figure's GIN row.

\textbf{The graph-network control sits at \textsc{n1}'s rung.} On tau-bench, \textsc{aware} beats
\textsc{n1} by $+0.051$ in the ladder of \secref{subsec:ladder}. A relation-aware graph
network separates the real layer from a count-preserving rewiring of it by $+0.047$ on the same
corpus (\appref{app:genericmisread}). \textsc{n1} and the network's rewiring agree in direction, and they
agree as two count-preserving controls sampled differently rather than as one intervention measured
twice: each holds only the per-graph edge count and neither holds per-step source degree, so the
network control sits at \textsc{n1}'s rung. Neither reads the degree-matched contrast.

Figure~\ref{fig:gnn} gives the per-corpus graph-network scores behind the summary in \secref{sec:misread}.

\subsection{Observed Dependency Span}
\label{app:span}

The saturation ratio $\rho$ (\secref{subsec:degenerate}) measures how many prior steps a layer attaches, not how far back they sit. To separate the two, for each observed dependency edge $u\to v$, where $u$ is the earlier source and $v$ the later step that uses it, we measure span $s(u,v)=\mathrm{idx}(v)-\mathrm{idx}(u)$ and report per corpus the median per-run span, the median per-run maximum span, the median normalized by $n-1$, and the share of edges with $s\ge 10$ (Table~\ref{tab:span}, six corpora, run under a fixed interpreter seed for reproducibility). The picture splits by edge type. The coding and web corpora are local at the median, with same-file or same-resource predecessor edges at span two to four, though OpenHands carries a delayed tail (median run maximum span $40$). tau-bench is not local: its writes attach to prior database reads in the same run, so the median span is $14.5$ and $76\%$ of edges span at least ten steps. All sit far below the ceiling, from $\rho=0.001$ to $\rho=0.018$, so span is a distinct axis from saturation. This is an access-derived span over recovered edges, not a gold causal reach: in tau-bench a write links to its conservative prior reads, and in the file corpora an edge links consecutive accesses to the same resource, so the number reports how local or delayed the recovered edges are, not a proven long-range reliance.

\begin{table*}[t]
\centering\small
\caption{Observed dependency span per corpus (per-run summaries over edgeful runs). Coding and web edges are local at the median; tau-bench read-to-write edges reach further. All sit between $\rho=0.001$ and $\rho=0.018$, so span is distinct from saturation. The span is access-derived over recovered edges, not gold reliance.}
\label{tab:span}
\begin{tabular}{@{}lcccc@{}}
\toprule
Corpus & med span & med$/n$ & med max & share $s\!\ge\!10$ \\
\midrule
tau-bench        & 14.5 & 0.47 & 22 & 0.76 \\
SWE-agent        & 2.0  & 0.07 & 4  & 0.10 \\
SWE-Gym          & 2.0  & 0.09 & 4  & 0.04 \\
OpenHands        & 4.0  & 0.03 & 40 & 0.37 \\
tau2-bench       & 3.5  & 0.18 & 6  & 0.08 \\
AgentRewardBench & 2.0  & 0.09 & 2  & 0.02 \\
\bottomrule
\end{tabular}
\end{table*}

\section{Localization Detail}
\label{app:loc}
\label{sec:execution}
\label{sec:localization}

\textbf{Over position baselines the execution layer improves top-3 and MRR localization but not top-1, an ancillary result on one corpus.} Where
the dependency layer predicts \emph{which} runs fail, the free layer ranks \emph{where} the fault
lies. On Who\&When, a multi-agent failure-attribution benchmark with step-level fault
labels~\cite{zhang2025whoandwhen}, one score learned over eight execution-layer features ranks the
126 failed runs' steps. Faults cluster (\FaultEarlyPct\% in the first third, \FaultCentralPct\% at the
most-active agent), so the comparison that matters is against an early-fault \emph{position prior}
rather than the random floor. The gain over that prior excludes zero on all three metrics under a
seed-block interval over five splits of the same runs. Resampling the 126 runs instead leaves top-3
and MRR clear of zero and puts top-1 at $[-0.002, +0.108]$, so the top-1 gain is the one that does
not survive a run-level interval. No ablation separates which of the
eight features carries the lift, and a cubic spline on position alone beats this model on top-1, so
the prior we measure against is a stated linear baseline.

\begin{table}[t]
\centering\small
\caption{Step-level fault localization on the Who\&When multi-agent failure-attribution
corpus (the 126 labeled-failed runs of its algorithm-generated subset, per-run mean over five
seeds). Each model ranks a failed run's steps so the labeled fault lands near the top.
Execution-graph structure beats the early-fault position prior, and both beat the random floor;
the structure-over-position gain has a seed-block 95\% interval excluding zero on all three
metrics, over five splits of the same runs; a 20{,}000-resample run-level bootstrap keeps top-3 and
MRR clear of zero and spans zero on top-1. The inferred dependency layer is a relabelling of
run order on this corpus (see text), so the lift is carried by the execution layer.}
\label{tab:localization}
\begin{tabular}{@{}lccc@{}}
\toprule
Ranking model & top-1 & top-3 & MRR \\
\midrule
Random (chance floor)    & 0.119 & 0.346 & 0.324 \\
Position prior           & 0.159 & 0.516 & 0.407 \\
\textbf{Execution structure} & \textbf{0.211} & \textbf{0.614} & \textbf{0.454} \\
\bottomrule
\end{tabular}
\end{table}

\subsection{Why the Dependency Layer Is Position-Equivalent Here}

The dependency layer does not help on Who\&When, and the reason is exact rather than incidental.
Who\&When records no per-step read or write, so its dependency layer is \emph{inferred} at the
weakest attachment grade: the full-history assumption that each step depended on every earlier
step. Under that assumption every dependency-structural feature of a step is a deterministic
function of the step's index and the run's length. For a step with zero-based index $i$ in an
$n$-step run, the number of earlier sources is $i$ and the number of later consumers is $n-1-i$,
and that earlier-source count normalized by $n-1$ is the position itself, on all $1{,}099$ steps of
the corpus; a linear ranker on that earlier-source count reproduces the position prior exactly in
all five seeds. The
layer is order in another notation, an instance of the degenerate regime in which a named
assumption collapses the dependency layer onto run order. That bounds what the layer encodes
rather than what any ranker reading it could score: a cubic spline on position alone beats the
linear prior on two of the three metrics, and beats the eight-feature execution model itself on
top-1 in four seeds of five and ties it in the fifth, so the prior is a stated baseline and the lift is measured against
that baseline rather than against the best reader of position. The attachment model anticipates the collapse. What a layer can do is
decided by the grade of its edges, not only by their presence: inferred at full history, the
dependency layer adds nothing to run order; observed, as in the coding corpora, it would not be
position-bound. The lift we measure is therefore an execution-layer gain over that stated
prior.

\subsection{The Limits of the Localization Result}

Position-equivalence bounds what the dependency layer carries on this corpus, and the
localization result reports what the execution layer carries in the same place. The execution
layer is the free, observed base: it defines the nodes and the run order. A reader could grant
all of that and still doubt that it carries failure information of its own. Localization
speaks to that doubt. On a corpus where the dependency layer is degenerate, a score over eight
execution-layer features puts the labeled faulting step above the position prior. The top-3
and MRR gains stay clear of zero under both the seed-block interval and the run-level
bootstrap. Measured against that stated linear prior, those gains are evidence that the
execution layer carries signal of its own. They do not say which kind of failure produced
them, since no experiment here sorts runs into coordination and reliance categories
(\appref{app:object}). So the base layer has a job of its own, and localization is it.

Three limits bound the claim. The result lives within a single corpus, so it is a faithfulness
result for the execution layer, not a transfer result; localization is reported run by run,
with no cross-corpus generalization claimed for it. It is also a layer-faithfulness ablation rather
than an entry on the Who\&When leaderboard: the trace-reading LLM attribution baselines in that
benchmark diagnose from the full transcript, while this test asks only whether execution topology
adds signal beyond time, so the two numbers answer different questions.
The mirror experiment for the dependency layer has not been evaluated here.
It requires step-level error annotations aligned with observed access events.
CodeTraceBench~\cite{li2026codetracer} provides access-bearing coding traces and step annotations,
making it a candidate subject to recovery and annotation-alignment checks.
Such a test would assess the dependency layer's localization value directly.

\section{An Exploratory Probe of Declared Workflow Graphs}
\label{app:declared}

FLORA-Bench~\cite{zhang2025flora} supplies workflow graphs over agents, together with a binary
pass/fail label per workflow-task pair, for workflows emitted by a planner (the G-Designer
subset) or by a program (the AFLOW subset). We use these graphs as workflow-level examples of
explicit dependency specifications. Their connections describe supplied workflow flow; they do
not establish which earlier result a particular execution step consumed, and $74$ of the $1051$
AFLOW HumanEval graphs are cyclic, so no edge-wise ordering $t(u) < t(v)$ holds across the
corpus. This probe therefore evaluates workflow summaries at agent granularity, while
validation of declared event-level reliance remains open.

Every group here carries the same declared designation, and the probe neither varies the
attachment grade nor supplies it to a learner. The AFLOW and G-Designer subsets differ in
emitter, workflow distribution and task sampling at once, so a difference between them
identifies none of those factors on its own.

We retain up to $4000$ rows per group from the first $120000$ rows of the train stream after
buffer shuffling with seed $0$ and buffer size $20000$. The eight retained groups hold $28006$
rows in total. Failure is the complement of the shipped success label. Node count is the number
of defined entries in the supplied workflow node dictionary, including isolated nodes and
excluding null fields introduced by struct materialization. Node counts vary within groups, so
the baseline uses both the node count $n$ and the count $m$ of distinct non-self directed
edges.

\begin{table*}[t]
\centering\small
\caption{Exploratory failure prediction from supplied FLORA-Bench~\cite{zhang2025flora}
workflow graphs, by task domain and emitter. \textsc{Size} is a standardized logistic model on
the number of defined nodes $n$ and distinct non-self directed edges $m$; {+}\,shape adds
$m/n$, normalized depth, and normalized maximum in- and out-degree. ROC-AUCs average five
stratified folds at each of five shuffle seeds within each group. The failure rate and the
median pair-normalized edge count $\rho = m/\binom{n}{2}$ describe the retained rows. Lifts are
differences of unrounded mean AUCs. The paired intervals are approximate corrected repeated-CV
diagnostics~\cite{nadeau2003inference}, without task or workflow cluster adjustment and without adjustment for selection
across the eight groups.}
\label{tab:flora}
\begin{tabular}{@{}llcccccc@{}}
\toprule
Domain & Emitter & Fail\% & $\rho$ & \textsc{Size} & {+}\,shape & Lift & Paired 95\% interval \\
\midrule
mbpp      & AFLOW      & 29 & 0.600 & 0.739 & 0.742 & $+0.003$ & $[-0.007, +0.013]$ \\
humaneval & AFLOW      & 56 & 0.361 & 0.567 & 0.662 & $+0.095$ & $[+0.050, +0.140]$ \\
mmlu      & AFLOW      & 53 & 0.400 & 0.621 & 0.691 & $+0.070$ & $[+0.055, +0.086]$ \\
gsm8k     & AFLOW      & 34 & 0.700 & 0.611 & 0.620 & $+0.010$ & $[-0.002, +0.021]$ \\
MATH      & G-Designer & 52 & 0.536 & 0.522 & 0.524 & $+0.002$ & $[-0.014, +0.018]$ \\
humaneval & G-Designer & 37 & 0.536 & 0.495 & 0.494 & $-0.001$ & $[-0.020, +0.018]$ \\
mbpp      & G-Designer & 52 & 0.600 & 0.497 & 0.492 & $-0.005$ & $[-0.019, +0.008]$ \\
mmlu      & G-Designer & 38 & 0.600 & 0.510 & 0.513 & $+0.004$ & $[-0.011, +0.018]$ \\
\bottomrule
\end{tabular}
\end{table*}

\textbf{Pair-normalized edge count.} The median of the eight group medians of
$m/\binom{n}{2}$ is $0.568$ (range $0.361$ to $0.700$). We report this as a descriptive
pair-normalized edge count for the supplied workflow graphs. These values do not support a
comparison with the observed layer, whose per-corpus medians run $0.001$ to $0.018$
(Figure~\ref{fig:degenerate}), and this statistic establishes neither independence from size nor
predictive utility. For cyclic workflow graphs $\binom{n}{2}$ is a numerical reference rather
than a full-history ceiling: a three-node directed cycle reaches $\rho = 1$ without being full
history.

\textbf{Two of eight intervals exclude zero.} In the sampled FLORA-Bench data, adding
four normalized workflow-graph summaries to a logistic baseline on defined node and edge counts
increases mean ROC-AUC by $+0.095$ on AFLOW HumanEval and $+0.070$ on AFLOW mmlu under repeated
within-group row-level cross-validation; these are the only two of eight groups whose
approximate paired corrected-CV $95\%$ intervals exclude zero (Table~\ref{tab:flora}). The
intervals for the other six groups include zero. These exploratory comparisons show predictive
gains for the specified workflow summaries in two AFLOW groups. They do not identify an effect
of attachment grade, establish a template mechanism, or establish absence of useful graph
information in the remaining groups. A declared designation alone does not establish predictive
utility; the measured increment is specific to the graph summaries, the learner, the baseline
and the evaluation population.

The approximate paired intervals describe the corrected repeated-CV comparison. They are
exploratory, they do not adjust for repeated tasks or workflows or for selection across the
eight groups, and they do not establish equivalence when they include zero. Adding $m/n$ to the
count baseline leaves increments of $+0.094$ on AFLOW HumanEval and $+0.066$ on AFLOW mmlu;
those comparisons address that additional count transformation, without establishing
independence from all size-based predictors.

The implemented depth feature is the longest directed path length for acyclic graphs and a zero
sentinel for the cyclic ones. Retaining that convention, the HumanEval increment is $+0.095$;
excluding the $74$ cyclic rows and refitting gives $0.575$ against $0.644$, an increment of
$+0.069$. That sensitivity analysis uses a different population and is reported separately, not
as the headline value.

\textbf{What this probe does and does not settle.} It reports a reproducible workflow-level
comparison on a public corpus, including its small and negative increments. It does not
instantiate the event-level recovery map of \secref{sec:representation}, because its graphs relate
agents rather than time-ordered events and some of them are cyclic. Empirical coverage of
declared event-level reliance therefore remains open, and a tool-call-level declared corpus with
step labels stays the cleaner target \appref{app:directions} names.

\section{Instrumentation and Open Directions}
\label{app:directions}
\label{sec:directions}

Instrumentation is the lever the rest of this agenda turns on. Logging each step's reads and writes
supplies the access events a raw trace omits, which is what carries an edge from inferred to
declared. The workflow-level probe of \appref{app:declared} does not evaluate such instrumentation,
and it does not establish a predictive gain from changing the grade. The same logs would build the
tool-call-level declared corpus that localizing a reliance fault needs, one carrying logged
dependencies and step-level fault labels together, which no benchmark supplies today. Whether the
marginal lift holds under genuine long-range reliance is untested, and richer web and embodied
settings are where the two-layer graph should be exercised next. Our first span audit
(\appref{app:span}) finds observed edges local in the coding corpora and longer where writes attach
to earlier reads, at the same low $\rho$. Pruning off-path reliance, repairing a run while it still
runs, and fleet-scale structural anomaly detection~\cite{xu2025fewshot} sit outside what this paper
demonstrates. Our unsupervised attempts at the last are weak and detector-inconsistent, and a static
dead-step scan tracks graph density rather than wasted work. For the graph networks of
\appref{app:gin}, a readout that uses the attachment grade is one design worth evaluating.

\section{Limitations in Detail}\label{app:limitsdetail}

\textbf{The transfer result holds under the fixed logistic probe.} This pattern is descriptive, with
no registered hypothesis, and the exploratory refit under a cubic spline and a gradient-boosted model
has no uncertainty interval (\appref{app:transferci}, \appref{app:d}). The registered primary rule
of the transfer ladder concerns another arm, size plus dependency against its degree-matched control.
Refit under the same two
learners, the dependency-only block falls below chance on two of the \NObsCorpora{} held-out corpora
under the gradient-boosted model and on four under the cubic spline, against one and three for size
plus dependency. The arm the abstract reports is therefore the more learner-sensitive of the two, and
the headline holds for the fixed logistic probe rather than for the feature block in general.
No transfer control rescales the raw size counts per corpus or builds the dependency-only block from a
counterfeit layer, so part of the transfer contrast may reflect feature scaling rather than recovered
dependencies.
\textbf{Two separate checks, never combined, bound the within-corpus lift, and one corpus carries it
through both.} Of the three corpora with a positive lift, tau-bench keeps it clear of zero under a
spline, a boosted tree, and task-grouped folds, though resampling tasks leaves its sign unresolved
(\appref{app:c1}, \appref{app:c2}). SWE-agent loses it
under task grouping. SWE-Gym keeps it under a boosted tree, reverses sign under a spline, and holds
one run per task, which leaves sensitivity to repeated-task leakage untestable there rather than
passed. Under flexible learners the increments are not separated from read and write composition
either: against an execution-side baseline with run-level folds, each logistic increment shrinks yet
stays positive in every seed, while each spline and boosted-tree point estimate is negative
(\appref{app:c1}). The corpus cards show that the tau-bench, SWE-agent and web corpora mix
generating models, and no analysis here separates an increment from model identity.
\textbf{The preregistered control did not separate the recovered attachment from a degree-matched
counterfeit within corpus; in transfer, only SWE-Gym clears Holm in the attachment-randomization
screen.} Two of the four features cannot move under that control (\secref{subsec:ladder}). That
screen gives nominal separations on \NTwoTrSeparates{} of the \NObsCorpora{} held-out corpora, and
the counterfeit scores higher on \NTwoTrInverts{}. The paired task-bootstrap contrasts clear Holm on
all \NTwoTrSeparates{} of those positive corpora within their separate six-corpus family
(\appref{app:n2transfer}). With run-size covariates and each step's earlier-source count held
fixed, that contrast measures sensitivity to replacing dependency endpoints; it does not apportion
absolute AUC between run size and dependency features (\secref{sec:unification}). Neither result
shows that the attachment grade is inert, and neither licenses a claim beyond the evaluated
extraction, feature probe, corpora and control (\secref{subsec:ladder}).

\textbf{Whole-family withholding yields no positive separation on any new target.} The attachment
screen has no Holm survivor among \FamNewTargets{} new targets. The paired task bootstrap has one among four eligible
targets: SWE-Gym, negative at $\FamSweGymDelta{}$. These \FamFamilies{} frozen task families support no
general claim across domains (\appref{app:familytransfer}).
\textbf{Coverage leans on coding and database tasks.} Five of the \NObsCorpora{} corpora are coding
or database runs; the web family holds one. \textbf{Horizon is short and long-range reliance is
untested.} The characterized runs are small at the median, about 10, 30, and 34 steps on
Who\&When, tau-bench, and SWE-agent, and the observed reliance available today is mostly local
(\appref{app:span}). Under long-range reliance, run size may become
a stronger baseline (\appref{app:directions}). \textbf{The observed grade is a principled
heuristic.} Its attachment rules are in \secref{sec:signal}; ground-truth reliance labels do not
exist at scale for any agent corpus. \textbf{The per-edge grade is a proposed schema.} No
experiment here mixes grades within a graph, so whether a model can use the per-edge stamp is
untested. \textbf{The graph-network comparison covers off-the-shelf
models only.} It does not establish which architectural change would close their transfer deficit
against the lean graph features (\appref{app:gin}). Open directions and undemonstrated uses are in
\appref{app:directions}.

\section{Related Work, Extended}
\label{app:related}

\textbf{Agent tracing and observability.} The execution layer is precisely what agent traces and
observability tooling already record~\cite{dong2024agentops}. A reason-act
trace~\cite{yao2023react} logs which step ran and in what order, and the public corpora behind our
results (\secref{sec:unification}) are themselves logged execution records. Recording the
execution layer is the free half: a trace gives which agent acted and how control moved, and stops
there. Even work that reconstructs an execution trace post hoc, when only final outputs
survive~\cite{nian2026implicit}, recovers the execution layer and not the reliance beneath it.
Within software engineering, one study puts the trajectories of three agents into a common
thought-action-result format and relates trajectory properties to
success~\cite{bouzenia2025understanding}; that format is shared within one setting and records each
run as a step sequence. What each step relied on is not recorded by the trace as a dependency edge. When access
events are absent, that reliance must be declared by instrumentation or inferred under an
assumption.
Observability that captures only execution therefore captures the base and misses the signal.

\textbf{Graph anomaly detection for agents.} A newer line builds an agent graph and scores
it for anomalies, almost always with a security framing: attacks, prompt injection, or malicious
agents~\cite{he2025sentinelagent,zhou2025guardian,wang2025gsafeguard}. How much of it is typed
varies: SentinelAgent distinguishes node and edge kinds, while GUARDIAN carries one of each. This
work shares the graph base but differs in aim. We evaluate one representation for faithfulness across
agent settings and make the dependency layer the object of study, rather than building a
single-objective security detector. Evidence that graph topology changes memory-leakage risk in
multi-agent systems~\cite{liu2026topology} is one reason to put structure at the center, and it
points the same detectors at reliability rather than only at attacks.

\textbf{Learned program graphs.} Software has been represented as a typed graph over code for
some time, and learned on. Code property graphs merge the syntax tree, control flow and data flow
into one typed multigraph for vulnerability discovery~\cite{yamaguchi2014cpg}; program graphs with
data-flow and lexical-use edges are learned with message passing for variable
tasks~\cite{allamanis2018learning}; and Devign learns over a composite of those
representations~\cite{zhou2019devign}. Two things separate this line from ours. Its graph is
static, recovered from source text rather than from a run, so an edge is a syntactic fact rather
than something a step did; and every edge arrives at one epistemic standing, with no rung for an
edge posited under an assumption. The two-layer object here is a record of an execution, and the
attachment grade is the axis those representations have no place for.

\textbf{Uncertain and probabilistic graphs.} Uncertain-graph and probabilistic-graph models
attach a scalar probability to each edge and reason about the distribution of graphs it
induces~\cite{potamias2010uncertain}. The attachment grade is related but finer along a different
axis. It does not say an edge is present with probability $p$; it records by what means the edge is
known to be present, observed from the trace, declared by instrumentation, or inferred under a
named assumption. Two edges can be equally certain yet differ in grade, and it is the grade, not a
probability, that tells a size-collinear inferred edge apart from a logged one
(\secref{sec:misread}). A single edge weight cannot represent that distinction.

\textbf{Temporal and dynamic heterogeneous graph learning.} Our object is a directed, typed,
temporal multigraph whose dependency edges point forward in time, which places it inside the dynamic heterogeneous graph family
that temporal and dynamic graph learning studies~\cite{kazemi2020dynamic}. Agent-side methods
already borrow from it: temporal graph modeling has been applied to multi-agent collaboration for
safety~\cite{zhou2025guardian}. What the general family does not contain is the degenerate regime
of \secref{sec:misread}. Because the dependency layer is sometimes inferred rather than
observed, a named full-history assumption collapses it to a deterministic function of run size,
with edge count $\binom{n_{\text{steps}}}{2}$ and a measured collinearity with run length as high
as \MaxInfDepSizeCorr{} on the evaluation corpora under that assumption and $1.00$ on the Who\&When runs of
Figure~\ref{fig:degenerate}. An edge weight says how strong an edge is and a temporal attribute says when
it arrived; neither says by what means it is known, so neither separates a logged reliance from one
a full-history assumption supplied. The
relation-aware models that do read the edge type, R-GCN and HGT, fall below chance on held-out
corpora (\appref{app:gin}), and the edge-type-blind network is fooled by the inferred layer exactly
as the linear probe is (\secref{subsec:gate}).

\textbf{Learning on small typed graphs.} Small graph size is not a general
barrier to message passing: molecular benchmarks such as MUTAG and QM9
support classification and property prediction on small labeled graphs
\cite{xu2019gin,ramakrishnan2014qm9,morris2020tudataset}.
This comparison does not identify why the graph networks underperform
the lean graph features on our agent runs. It does show that small size and
typed nodes alone do not distinguish agent graphs from established
graph-learning settings.

\textbf{Edge ablations on agent graphs.} Concurrent work ablates edge types under a
different construction: holding turn nodes and a two-layer GCN fixed, AFANet's single-run point
estimates put adjacent-turn edges alone at 74.82 agent micro-F1 on AEGIS-Bench against 74.16 for
adjacent-turn and same-agent edges together, so the richer edge set does not win every metric,
although its authors conclude that both edge types contribute \cite{li2026afanet}.
\citet{shi2018tree} likewise found trivial trees competitive with real parse trees
across ten tasks when only the structure-generating rule changed.

\section{AI Use Statement}
\label{app:ai-use}

The author used AI coding and writing assistants to support implementation, analysis tooling, figure preparation, and manuscript drafting and revision. The reported empirical results are computed by evaluation code over benchmark records and saved experimental artifacts. The author takes full responsibility for the methods, interpretation, and final content of the paper.

\end{document}